\documentclass[10pt,journal,compsoc]{IEEEtran}
\usepackage{pbox}
\usepackage{amsmath,lipsum}
\usepackage{amssymb}
\usepackage{tikz}

\usepackage{algorithmicx}
\usepackage{algorithm}% http://ctan.org/pkg/algorithms
\usepackage{algpseudocode}% http://ctan.org/pkg/algorithmicx
\usepackage[linesnumbered,ruled,vlined,boxed,algo2e]{algorithm2e}
\usepackage{mathtools}

\usepackage{commath}
\usepackage{subfigure}
\usepackage{multirow}
\usepackage{makecell}
\usepackage{booktabs} % for professional tables
\usepackage{accents}
\usepackage{bm}
\usepackage{ulem} % for strike through

\newcommand{\Te}{Teacher}

\newcommand{\St}{Student}

\usepackage{bm}
\usepackage{multirow}
\usepackage{pifont}
\newcommand{\cmark}{\ding{51}}
\newcommand{\xmark}{\ding{55}}

\ifCLASSOPTIONcompsoc
  \usepackage[nocompress]{cite}
\else
  \usepackage{cite}
\fi
\ifCLASSINFOpdf
\else
\fi
\usepackage{amsmath}
\begin{document}
\title{Mining \textit{Data Impressions} from Deep Models as Substitute for the Unavailable Training Data}
%\title{Data Impressions: Mining Deep Models to Extract Surrogate Samples for Substituting Unavailable Training Data}


\author{Gaurav Kumar Nayak,~\IEEEmembership{Graduate Student Member,~IEEE,}
        Konda Reddy Mopuri, Saksham Jain,
        and Anirban~Chakraborty,~\IEEEmembership{Member,~IEEE}% <-this % stops a space
%\IEEEcompsocitemizethanks{\IEEEcompsocthanksitem K. R. Mopuri is with the Department of Computer Science and Engineering, Indian Institute of Technology Tirupati}
%of Electrical and Computer Engineering, Georgia Institute of Technology, Atlanta,
%GA, 30332.\protect\\
%E-mail: see http://www.michaelshell.org/contact.html
%\IEEEcompsocthanksitem J. Doe and J. Doe are with Anonymous University.}% <-this % stops an unwanted space
%\thanks{Rest of the authors are with the Department of Computational and Data Sciences, Indian Institute of Science, Bangalore, India, 560012. E-mail: gauravnayak@iisc.ac.in, kmopuri@iittp.ac.in, sakshamjain@iisc.ac.in, anirban@iisc.ac.in.}}

\IEEEcompsocitemizethanks{\IEEEcompsocthanksitem G. K. Nayak and A. Chakraborty are with the Department of Computational and Data Sciences, Indian Institute of Science, Bangalore, India.
\IEEEcompsocthanksitem S. Jain is currently with the Department of Electrical and Computer Engineering, Duke University USA. He was affiliated with the Department of Computational and Data Sciences, Indian Institute of Science when this work was carried out.
\IEEEcompsocthanksitem K. R. Mopuri is with the Department of Computer Science and Engineering, Indian Institute of Technology Tirupati.}
\thanks{For all correspondence: Anirban Chakraborty (anirban@iisc.ac.in)}}


\markboth{Journal of \LaTeX\ Class Files,~Vol.~14, No.~8, August~2015}%
{Shell \MakeLowercase{\textit{et al.}}: Bare Demo of IEEEtran.cls for Computer Society Journals}



\IEEEtitleabstractindextext{%
\begin{abstract}
Pretrained deep models hold their learnt knowledge in the form of model parameters. These parameters act as ``memory" for the trained models and help them generalize well on unseen data. However, in absence of training data, the utility of a trained model is merely limited to either inference or better initialization towards a target task. In this paper, we go further and extract synthetic data by leveraging the learnt model parameters. We dub them \textit{Data Impressions}, which act as proxy to the training data and can be used to realize a variety of tasks. These are useful in scenarios where only the pretrained models are available and the training data is not shared (e.g., due to privacy or sensitivity concerns). We show the applicability of data impressions in solving several computer vision tasks such as unsupervised domain adaptation, continual learning as well as knowledge distillation. We also study the adversarial robustness of lightweight models trained via knowledge distillation using these data impressions. Further, we demonstrate the efficacy of data impressions in generating data-free Universal Adversarial Perturbations (UAPs) with better fooling rates. Extensive experiments performed on benchmark datasets demonstrate competitive performance achieved using data impressions in absence of original training data.
\end{abstract}

}

\maketitle

\IEEEdisplaynontitleabstractindextext

% \fi
\IEEEpeerreviewmaketitle

\IEEEraisesectionheading{\section{Introduction}
\label{sec:introduction}}

%\input{intro}
\IEEEPARstart{S}{upervised} learning typically requires large volumes of labelled data. Training of sophisticated deep neural networks (DNNs) often involves learning from thousands (MNIST~\cite{lecun1998gradient}, CIFAR~\cite{krizhevsky2009learning}) (sometimes millions, e.g. ImageNet~\cite{imagenet-ijcv-2015}) of data samples.
\iffalse %%%%%%%commenting starts
Resulting trained models hold the task specific knowledge obtained from these samples in the form of their parameters. Often this knowledge is transferred for solving related tasks in the form of pre-training. That is, subset of early layers (leaving the task specific ones) learned by the models are taken to build other models with freshly initialized target specific layers on top. In other words, parameters of the trained models act as better initialization for learning the target model's parameters. This process often results in advantages such as faster convergence~\cite{rethinking-iccv-2019} and better generalization (e.g.~\cite{faster-nips-2015}) in case of low target data scenarios. Also, multiple works (e.g.~\cite{taskonomy-cvpr-2018}) have studied the affinity between pairs of tasks for better transfer learning in computer vision. Therefore pre-training has been established as a way of utilizing the knowledge from a dataset in the form of learned parameters.
\fi %%%%%%%commenting ends
Despite their ability to train complex models, these training datasets pose practical challenges. These datasets (i) are often huge in size (e.g. ImageNet~\cite{imagenet-ijcv-2015}), (ii) are proprietary, and (iii) involve privacy concerns (e.g. biometric, healthcare data). Hence, in practice, public access to the data samples used for training may not always be feasible. Instead, the resulting trained models can be made available relatively easily. For instance, Facebook's Deepface~\cite{deepface-cvpr-2014} model is trained over $4$M confidential face images.

However, in the absence of training data, a trained model has limited utility for adapting it to a related task. In this scenario, the best thing one can do is utilizing the trained layers as a better initialization for a fresh training. In other words, unavailability of the training data restricts the transfer learning possibilities to a mere pretraining. Because of this, applications with more practical significance such as Knowledge Distillation (KD) and Unsupervised Domain Adaptation can not be realised in the absence of the training data. For instance, in the KD framework, to compress a sophisticated (Teacher) Neural Network into a light weight (Student) one, training data is required as the transfer set. Matching the input output behaviour of the models (despite their architectural differences) which is the key for model compression can not take place in the absence of training data. Given no prior information about the underlying training data, it is challenging to compose a suitable transfer set to replace it. Similarly, for Unsupervised Domain Adaptation, data on which the source model is trained plays a vital role for an effective knowledge transfer. In most target scenarios, nontrivial performances can be achieved by suitably adapting the source models. All these possibilities get abolished when we do not have the training data along with the trained model. This leads to a massive under utilization of the training efforts. Therefore, in this paper we investigate for approaches that can craft proxy data for extending the utility of a trained model beyond pretraining. Specifically, we propose to utilize the given trained model itself for extracting the proxy data. %{\color{red} something} In this context, we investigate the following questions: \textit{Is it possible to extract the data samples that trained a given deep neural network? Can we use these extracted samples in place of original data to train other models?} 

%In this paper, different from pre-training, we propose a method for extracting knowledge from the trained deep networks in the form of data samples. Further, we demonstrate that the extracted synthetic data can act as proxy for the training data. 

We consider the Convolutional Neural networks (CNNs) trained for object recognition. Starting from Knowledge Distillation, we explore data-free adaptation of the trained model in various application scenarios. Inspired from Mopuri \textit{et al.}~\cite{aaa-eccv-2018}, we extract impressions of training data from the parameters of the trained CNN model. Note that with no external prior about the training data, we perform the proxy data synthesis required for the adaptation. We extract the inter-class similarities from the CNN parameters and model the output (softmax) space of the classifier using a family of Dirichlet distributions. We sample these distributions and iteratively reconstruct the corresponding data samples in the input space from random initializations. Our approach extracts the proxy data samples from a trained model one at a time. The recovered synthetic data samples are named \textit{Data Impressions} (DIs) as they are the impressions of actual data extracted from the model. Note that the extraction process neither requires original training data nor any prior information, and the extracted samples act as a proxy in the absence of original data. Importantly, the extraction of the impressions is agnostic to the downstream application. In other words, the same method can be applied directly across multiple applications. This observation denotes that they capture generic knowledge about the training dataset suitable for adapting to various application.

One way to ensure the effectiveness of the extracted surrogate samples is via generalization. That is, by demonstrating that the extracted samples can be reliably used for adapting the model and generalize well on to the actual test data. Hence, for each adaptation we empirically verify the performance of the adapted models on the actual test datasets. In order to show the effectiveness of such generated data impressions, we leverage several computer vision applications that have faced problems arising from data-free set up. These problems have been tackled independently in the literature and various methods have already been proposed. We simply leverage these problems and propose solution strategies utilizing the aforementioned data impressions. We observe strong performances across all these tasks, thereby proving the utility of our data impressions as surrogates to the original training data. 

Here we would like to emphasize that these applications are to demonstrate the effectiveness of the data impressions and prove that they are reliable surrogates for the original training data samples. Hence it may be unfair to compare the performance with the corresponding dedicated data-free solutions for the individual applications. Also, given the generic nature of the data impressions, they may be utilized in several other tasks apart from the ones that we discuss in this work. %Thats how the application of this paper goes beyond the task at hand that we show.

The overall contributions of our work can be listed as follows:
\begin{itemize}
    \item We propose the first and generic framework for data-free adaptation of trained neural networks via extracting proxy data samples, called `Data Impressions'. We achieve this with no additional prior about the training data and without requiring any metadata about the resulting feature distribution.
    \item We study the  the extensive applicability of Data Impressions towards multiple applications such as Knowledge Distillation, Unsupervised Domain Adaptation, crafting Adversarial Perturbations, and Incremental Learning. We show that in the absence of original training data, Data Impressions can successfully train models that generalize well onto the actual test data.
    \item  Further, we study the robustness properties of the student models trained on the Data Impressions against adversarial perturbations. Experimental results demonstrate that Data Impressions consistently uphold the robustness properties of the corresponding teachers.
\end{itemize}
\iffalse
This work extends our earlier conference paper~\cite{zskd-icml-2019} and makes the following overall contributions: %mention the last two bullet
\begin{itemize}
    \item We proposed ({originally introduced in~\cite{zskd-icml-2019}}) the first and generic framework, called Data Impressions for data-free adaptation of trained neural network models via extracting proxy data samples. We achieve this with no additional prior about the training data distribution and without requiring any meta data about the resulting feature distribution.
    \item {Additionally, in this paper} we study the extensive applicability of Data Impressions towards multiple applications such as Unsupervised Domain Adaptation, crafting Adversarial Perturbations, and Incremental Learning in addition to Knowledge Distillation. We demonstrate that in the absence of original training data, Data Impressions can successfully train models that generalize well onto the actual test data. %Excellent generalization achieved by these models on the test sets composed of actual data samples reinforce that our Data Impressions capture essential properties of training data from the trained CNN classifiers.
    \item We further study the adversarial robustness properties of the student models trained on our Data Impressions and demonstrate that they consistently uphold the robustness properties of the corresponding teachers.
\end{itemize}
\fi 

Note that the framework for extracting the Data Impressions and their application Zero-Shot Knowledge Distillation were originally introduced in our earlier conference paper~\cite{zskd-icml-2019}. All the other contributions are novel additions to this extended article.

% highlight the last two bullets as the novel contributions
The rest of this paper is organised as follows: section~\ref{sec:related-works} discusses the existing works that are related to this research, section~\ref{sec:proposed-approach} presents our approach for extracting the Data Impressions from a trained CNN classifier, section~\ref{sec:applications} demonstrates the effectiveness of the approach via learning multiple related tasks, section~\ref{sec:overall_discussion} discusses the major findings across experiments on different applications and finally section~\ref{sec:concluion} summarizes the paper with conclusions.
%\vspace{-0.16in}
\section{Related Work}
\label{sec:related-works}
Our work introduces a novel problem of restoring training data from a trained deep model. It is broadly related to visualization works such as~\cite{backprop-iclrw-2014,guidedbackprop-iclrw-2015}. However, the general objective driving visualization works is to identify the patterns in the stimuli for which the neurons maximally respond and thereby alleviate their black-box nature. Based on the gradient driven visualization ideas, Mopuri~\textit{et al.}~\cite{aaa-eccv-2018} craft class representative samples, known as Class Impressions, from a trained CNN based classifier. Their objective is specific, which is to use these impressions for crafting adversarial perturbations in a data-free scenario. We extend this idea and make it a generic problem of extracting the samples that can substitute the training data. %\sout{We realize this via modeling the classifier's output space as a mixture of Dirichlet distributions and mapping the samples from it to input space.} 
Further, we demonstrate the effectiveness of our Data Impressions by successfully learning diverse set of related tasks over the restored data. Specifically we perform Knowledge Distillation, UAP (Universal Adversarial Perturbation) generation, Domain Adaptation, and Incremental Learning. For ease of reference, we briefly introduce these tasks and compare our idea of using Data Impressions with the corresponding existing works.

\textbf{Knowledge distillation}: is a process of emulating a large model called \textit{Teacher} by a lightweight model called \textit{Student}. The teacher model generally has high complexity and is not preferred for real-time embedded platforms due to its large memory and computational requirements. In practice, networks which are compact and lightweight are preferred. Existing works use training data (e.g.~\cite{hinton2015distilling,bucilua2006model}) or meta data (e.g.~\cite{dfkd-nips-lld-17}) extracted from the teacher for performing distillation. However, the proposed method transfers the knowledge without using either of them.
\iffalse
We can categorize several variations in knowledge distillation broadly into three types based on the amount of data used:

\textbf{Using entire training data or similar data}: In \cite{bucilua2006model}, model compression technique is used. The target network is trained using the pseudo labels obtained from the larger model with an objective to match the pre-softmax values (called logits). In \cite{hinton2015distilling}, the softmax distribution of classes produced by teacher model using high temperature in its softmax (called ``soft targets") are used to train the student model as the knowledge contained in incorrect class probabilities tends to capture the teacher generalization ability better in comparison to hard labels. The matching of logits is a special case of this general method. In \cite{furlanello2018born}, knowledge transfer is done across several generations where the student of current generation learns from its previous generation. The final predictions are made from the ensemble of student models using the mean of the predictions from each student. 

\textbf{Using few samples of original data}: In \cite{kimura2018few}, knowledge distillation is performed using few original samples of training data which are augmented by ``pseudo training examples". These pseudo examples are obtained using inducing point \cite{snelson2006sparse} method via iterative optimization technique in an adversarial manner which makes the training procedure complicated.
\fi
%%%%%%%%%%%%
To the best of our knowledge, our work (Nayak~\textit{et al.}~\cite{zskd-icml-2019}) is the first to demonstrate knowledge distillation in case where no training data is available. Contemporary to our work, Chen~\textit{et al.}~\cite{chen2019data} Micaelli \textit{et al.}~\cite{micaelli2019zero} and Addepalli \textit{et al.}~\cite{degan-aaai-2020} also attempt to perform knowledge transfer in the data-free scenario. However, unlike our activation maximization approach, they train GAN-inspired generative models to learn the proxy or fake data samples required for the transfer. These methods train the GAN with multiple objectives to ensure learning (i) difficult pseudo (or proxy) samples on which the Teacher and Student differ (\cite{micaelli2019zero}), (ii) uniform distributions over the underlying classes (\cite{degan-aaai-2020}), and (iii) samples predicted with a strong confidence by the Teacher model, (\cite{chen2019data}) etc. so that the transfer performance is maximized. Note that~\cite{degan-aaai-2020} uses arbitrary but natural proxy data for transferring the knowledge. Another generative model known as KegNet~\cite{knowledge-extraction-neurips-2019} by Yoo~\textit{et al.} also employs a conditional GAN framework along with a decoder objective for encouraging diversity in the fake images used for knowledge transfer. Unlike these GAN based approaches, our method do not involve such complex training procedures and do not require any ``proxy" data samples as used in~\cite{degan-aaai-2020}, thereby strictly adhering to the ``zero-shot" paradigm. 

On the other hand, recent works by Yin \textit{et al.}~\cite{dreaming-to-distill-cvpr-2020} and Haroush~\textit{et al.}~\cite{knowledge-within-cvpr-2020} attempt to synthesize the class conditional samples from a trained neural network model and enforce intuitive priors to improve the quality of the generated samples. Specifically they utilize the Batch Normalization (BN) layers' statistics such as feature mean and covariances extracted from the Teacher as the useful prior while synthesizing via maximizing the logit activations. Further, \cite{dreaming-to-distill-cvpr-2020} also imposes natural image priors such as smoothness while synthesizing the pseudo samples. Similarly Shoukai~\textit{et al.}~\cite{generative-low-eccv-2020} present a conditional GAN framework for quantizing a trained Teacher model in the data-free scenario by learning fake images. Their method along with utilizing the Batch Norm statistics for matching the training data distribution, also uses the Knowledge Distillation and CrossEntropy objectives for compressing the Teacher model. That way these works can be thought of improvements to our method but restricted to invert the models that have BN layers. These methods by design, are restricted to specific network architectures that use the batchnorm layers and hence cannot be utilized for older architectures or recent models that do not include such layers. Our framework, on the other hand, are completely independent of the pretrained network architecture and hence are more widely applicable. %Thus, a vast majority of the existing works for performing data-free knowledge distillation are generative (GAN based) in nature. whereas the proposed data impressions is a simple activation maximization framework that crafts the pseudo sample one at a time. 
Also, additionally in this work we perform robustness study on student models trained in the data-free setup.
%\textbf{check}

%It has been shown by \cite{mopuri2018ask} that the pretrained models have memory in terms of learned parameters and can be used to extract class representative samples. Although, it was used in the context of adversarial perturbation task, we argued that carefully synthesized samples can be used as pseudo training data for knowledge distillation.

\textbf{Incremental Learning}: Here the goal is to train a model using the samples of new classes without forgetting the knowledge gained from the old classes. With the limited memory constraints, several rehearsal based methods such as Rebuffi~\textit{et al.}~\cite{ rebuffi2017icarl} and Castro \textit{et al.}~\cite{castro2018end}, carefully store few samples from the old classes to avoid catastrophic forgetting. Pseudo-rehearsal methods like Shin~\textit{et al.}~\cite{shin2017continual} avoid storing samples from old classes, but instead they learn a generator which is trained using old class data. %At each incremental step, the model is trained with samples from the generator and the new class data, followed by retraining of the generator with the generated data from the previous step and the samples from new classes. 
Thus, there is an implicit assumption of the availability of the trained generator which is as good as having access to old class data. Moreover training of a generator has its own difficulty like mode collapse which requires proper handling. 

In cases where samples belonging to old classes are unavailable and only have access to the pretrained model which is trained on those classes, above discussed methods perform poorly. LwF~\cite{li2017learning} by Li \textit{et al.} tries to overcome this problem by only utilizing samples belonging to the new classes. The model is trained with these samples where cross entropy loss is used on new classes while distillation is applied on old classes.
%Here the goal is to train a model using the samples of new classes without forgetting the knowledge gained from the old classes. With the limited memory constraints, few works such as Rebuffi~\textit{et al.}~\cite{ rebuffi2017icarl} and Castro \textit{et al.}~\cite{castro2018end} carefully store few samples from the old classes to avoid catastrophic forgetting. Shin~\textit{et al.}~\cite{shin2017continual} avoids storing samples from old classes, but instead they learn a generator which is trained using old class data. But in cases where samples belonging to old classes are unavailable and only the pretrained model trained on old classes is available, such methods perform poorly. LwF~\cite{li2017learning} by Li \textit{et al.} only utilizes new class data where cross entropy loss is used on new classes while distillation is applied on old classes.
Recently, DMC~\cite{zhang2020class} by Zhang \textit{et al.} has shown great results by using auxiliary data and dual distillation loss. Their performance is dependent on how close the selected auxiliary data is to the training data distribution. Often domain knowledge is required for a careful selection of such data, which becomes a non-trivial task when no prior on the original training data is available except for a model trained with old classes. We, on the other hand, do not use any auxiliary data but instead generate Data Impressions from the model trained on old classes and use them with new class data to train the combined model. %We perform significantly better than LwF and comparable to DMC without any explicit use of any auxiliary data.

\textbf{Unsupervised Domain Adaptation}: The goal in this task is to adapt the model trained on source data to predict the labels for the unlabelled target data. Most of the existing works such as~\cite{ganin2015unsupervised, tzeng2017adversarial, tzeng2014deep, gretton2006kernel, lee2019drop} depend on the availability of both source and target data to perform the adaptation. However, recent work by Kundu~\textit{et al.}~\cite {kundu2020universal} overcomes this limitation, but only in the deployment stage. In the procurement stage, they require the source model to be trained not only on source training samples but also on negative source samples simulated with the help of source data. %Thus, their adaptation method in the deployment stage can't be applied on any given pretrained source model unless the pretrained model is equipped to handle out-of-distribution samples. 
As a main difference to ~\cite {kundu2020universal}, we restrict ourselves to closed set domain adaptation and we leverage Data Impressions in the absence of source data to perform source-free domain adaptation. %{In summary,~\cite {kundu2020universal} needs to deal with out-of-distribution (OOD), so they need special training procedure, while our method doesn't need to, because we assume we don't need to deal with OOD.} 
Please note that under this setup, our method is generic and can be applied on any trained source model.

Recently, Liang \textit{et al.}~\cite{liang2020we} proposed a new method (SHOT), which aligns the target features to the source hypothesis in the absence of the source data and target labels. The source-trained model is composed of feature extractor and classifier. The classifier module (hypothesis) is frozen and the feature encoding module is finetuned to fit the hypothesis using information maximization loss along with pseudo labeling via self supervision. Even though they obtain promising results, their method is heavily dependent on the architecture of the source network. They require weight normalization in the fully connected layer of the classifier and batch normalization layer at the end of the feature extractor module. %Also, the source model needs to be trained with label smoothing technique. 
The performance on the target data drops significantly when such dependencies are not met. On the other hand, our method does not have any such architectural dependencies.

Another recent work by Kurmi \textit{et al.}~\cite{kurmi2021domain} proposed an end to end framework where generation of pseudo samples and their use for adaptation in the absence of source data are performed simultaneously. In the generation module of~\cite{kurmi2021domain}, samples are synthesized using conditional GAN by modelling the joint distribution of images and corresponding labels of the source data. Additionally the authors use an adversarial discriminator to close the gap between the distribution of generated samples to that of %enforce the distribution of generated samples to be close to the distribution of 
target samples. So, they require target data in order to generate pseudo-samples. Also, their overall loss function is a combination of multiple losses which requires careful balancing. Unlike theirs, our generation method is independent of the target data. We generate data impressions using only the pretrained source model and is generic as its synthesis does not depend on the target data. Moreover, we do not perform any complicated GAN training. As our synthesis of samples is done using a single loss function, the optimization is easy to handle. 
%Our method utilizes Data Impressions which acts as a substitute in the absence of source data. These impressions are obtained from the pretrained source model and is generic as its synthesis is independent of the target data.} %Our method is generic and 
%Our generated impressions can then be used along with unlabelled target data to adapt any trained source model. %which allows any domain adaptation technique to be used on top of it.

\textbf{Universal Adversarial Perturbations (UAPs)}:  UAPs or Image agnostic adversarial perturbations are structured, mild noises that upon adding to the data, can confuse deep classifiers and enforce them to predict incorrectly. The training data (e.g. Moosavi-Dezfooli~\cite{moosavi2017universal}) is generally required to craft the UAP. Mopuri \textit{et al.}~\cite{fff-bmvc-2017,mopuri2018generalizable}, for the first time, presented a data-free approach for crafting UAPs using an activation maximization objective. Later they proposed Class Impressions~\cite{mopuri2018ask} as a way to reduce the gap between data-free and data-driven approaches. Proposed Data Impressions capture the patterns from the training data better than their class impressions and thereby can craft UAPs with better fooling ability.
%{\color{red} Gaurav will add something and make it smooth.}

%{\color{blue}
\textbf{Summary of differences with data-free methods}: %In summary, 
Several methods such as \cite{micaelli2019zero, chen2019data, knowledge-extraction-neurips-2019, knowledge-within-cvpr-2020, generative-low-eccv-2020, kundu2020universal, liang2020we, kurmi2021domain, mopuri2018ask, zhang2020class} have been proposed in the data-free set up towards different applications which are specifically designed. However, such methods are dedicated data-free solutions for individual applications. Hence, they are application specific where the data generation process is tied to the task at hand. %However, they are application specific where they generate samples only towards the task at hand. 
On the other hand, our proposed data impressions are synthesized without considering any downstream target task. We evaluate their efficacy by exploring their applications on different downstream tasks. %It can be noted that despite having an application agnostic generation process as well as using the same set of generated samples across a diverse set of data-free machine learning tasks, we observed very respectable performance throughout, unlike these task specific methods which are designed for particular tasks and cannot be applied across other applications.Even then, we evaluate their performance on multiple applications. 
We demonstrate that such impressions are indeed true substitutes of original training data samples and are suitable to be utilized across different applications.

Recently, Yin \textit{et al.}~\cite{dreaming-to-distill-cvpr-2020} also shows the utility of their pseudo samples on data-free pruning and continual learning besides their application in knowledge distillation. Their method \textit{`Deep Inversion'} is an extension of \textit{`Deep Dream'}~\cite{mordvintsev2015inceptionism} where they additionally regularize the feature distribution of generated data by matching the batchnorm statistics. Their method assumes the presence of batchnorm layers which are prevalent only in the modern networks. Hence, the performance of their method is heavily dependent on the number of batch norm layers in the intermediate layers of the trained classifier. %In situations where very few or no batchnorm layers are present in the classifier network, their method would yield similar low performance as~\cite{mordvintsev2015inceptionism}. 
They further boost their performance by  an iterative method \textit{`Adaptive DeepInversion'} that generates samples which cause teacher-student disagreement. As the student is involved in the loop, this scheme is application dependent and is also very similar to~\cite{micaelli2019zero}. % which uses kl divergence instead of Jensen-Shanon divergence. 
Their overall loss optimization contains a sum of many regularization losses, where finding appropriate weightage of the individual losses is troublesome. On the other hand, our generation strategy of data impressions does not depend on batchnorm layers in the trained classifier. This makes our framework independent of the pretrained network architecture and hence is more widely applicable. In other words, % nor involve any additional hyperparameter apart from learning rate. Hence our generation procedure involves simple optimization and is more generic as 
our method is not only application-independent but also architecture-independent. Apart from data-free knowledge distillation, we also show the utility of our generated impressions on a diverse set of applications which are disjoint in comparison to~\cite{dreaming-to-distill-cvpr-2020} (such as source-free unsupervised domain adaptation, and data-free universal adversarial perturbations). Moreover, for the first time we study the robustness properties of a student distilled in a data-free scenario.

We now discuss in detail our proposed approach for synthesizing data impressions.
\section{Proposed Approach: Extracting Data Impressions From Trained Models}
\label{sec:proposed-approach}
%\emph{Here, we present the proposed idea of extracting data from a trained model. We explain only the DIs part. We plan to present the possible applications in the next section. Most of the content here can be copied from our ICML paper.}

In this section we describe the proposed method to extract samples from a \textit{Trained} neural network model, which can act as substitute to the original training data. We first model the output (softmax) space of the \textit{Trained} classifier using a probability distribution. Then we sample softmax vectors from this distribution. For each softmax vector, we generate corresponding input via iteratively updating a random input. Modelling of the softmax space and estimation of the distribution parameters is explained in sec.~\ref{subsec:dirichlet-modelling} while the procedure to generate the samples from the sampled softmax vectors is described in sec.~\ref{subsec:crafting}
\vspace{-0.1in}
\subsection{Modelling the Data in Softmax Space} %Data Impressions as the transfer set
\label{subsec:dirichlet-modelling}
In this work, we deal with the scenario where we have no access to (i) any training data samples, or (ii) meta-data extracted from it~(e.g. \cite{dfkd-nips-lld-17}). In order to tackle this, our approach taps the memory (learned parameters) of the \textit{Trained} model and synthesizes pseudo samples to represent the underlying data distribution on which it is trained. Since these are the impressions of the training data extracted from the trained model, we name these synthesized input representations as Data Impressions. We argue that these can serve as effective surrogates for the training samples, which can be used to perform several tasks such as knowledge distillation, incremental learning, and unsupervised domain adaptation.

In order to craft the Data Impressions, we model the output (softmax) space of the \textit{Trained} model. Let $\bm{s}\sim p(\bm{s})$, be the random vector that represents the neural softmax outputs of the \textit{Trained} model, %$T(x,\theta_T)$ 
$T$ with parameters $\theta_T$. We model $p(\bm{s}^k)$ belonging to each class $k$, using a Dirichlet distribution which is a distribution over vectors whose components are in $[0,1]$ range and their sum is $1$. Thus, the distribution to represent the softmax outputs  $\bm{s}^k$ of class $k$ would be modelled as, $Dir(K, \bm{\alpha}^k )$, where $k \in \{1 \ldots K\}$ is the class index, $K$ is the dimension of the output probability vector (number of categories in the recognition problem), and $\bm{\alpha}^k$ is the concentration parameter of the distribution modelling class $k$. The concentration parameter $\bm{\alpha}^k$ is a $K$ dimensional positive real vector, i.e, $\bm{\alpha}^k$ $= [\alpha^k_1,\alpha^k_2,\ldots,\alpha^k_K ], \text{and} ~ \alpha^k_i >0, \forall i \in \{1, 2, \ldots K\}$.

\textbf{Concentration Parameter ($\bm{\alpha}$)}: Since the sample space of the Dirichlet distribution is interpreted as a discrete probability distribution (over the labels), intuitively, the concentration parameter $(\bm{\alpha})$ can be thought of as determining how ``concentrated" the probability mass of a sample from a Dirichlet distribution is likely to be. With a value much less than $1$, the mass will be highly concentrated in only a few components, and all the rest will have almost zero mass. On the other hand, with a value much greater than $1$, the mass will be dispersed almost equally among all the components.

Obtaining prior information for the concentration parameter is not straightforward. The parameter cannot be the same for all components since this results in all sets of probabilities being equally likely, which is not a realistic scenario. For instance, in case of CIFAR-$10$ dataset, it would not be meaningful to have a softmax output in which the dog class and plane class have the same confidence (since they are visually dissimilar). Also, same $\alpha_i$ values denote the lack of any prior information to favour one component of sampled softmax vector over the other. Hence, the concentration parameters should be assigned in order to reflect the similarities across the components in the softmax vector. Since these components denote the underlying categories in the recognition problem, $\bm{\alpha}$ should reflect the visual similarities among them.

Thus, we resort to the \textit{Trained} network for extracting this information. We compute a normalized class similarity matrix $(C)$ using the weights $W$ connecting the final (softmax) and the pre-final layers. The element $C(i,j)$ of this matrix denotes the visual similarity between the categories $i$ and $j$ in $[0,1]$. Thus, a row $\bm{c}_k$ of the class similarity matrix $(C)$ gives the similarity of class $k$ with each of the $K$ categories (including itself). Each row $\bm{c}_k$ can be treated as the concentration parameter $({\bm\alpha})$ of the Dirichlet distribution $(Dir)$, which models the distribution of output probability vectors belonging to class $k$.

\begin{figure}[htp]
  \centering
  \includegraphics[width=0.48\textwidth]{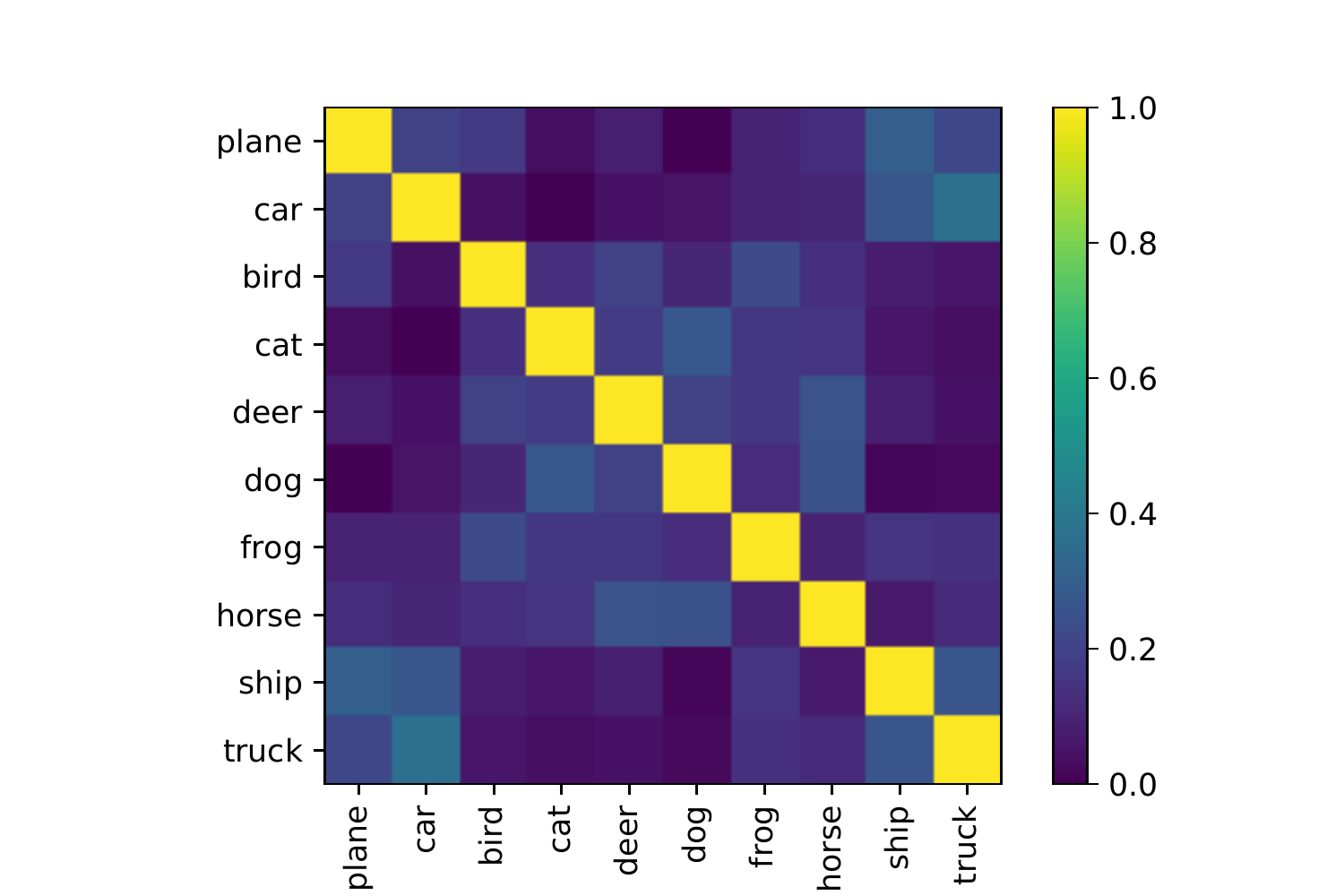}%
  \caption{Class similarity matrix computed for the \Te{} model trained over CIFAR-$10$ dataset. Note that the class labels are mentioned and the learned similarities are meaningful.}
  \label{fig:similarity-matrix-cifar}
  \end{figure}
  
\textbf{Class Similarity Matrix}: The class similarity matrix $C$ is calculated as follows. The final layer of a typical recognition model will be a fully connected layer with a softmax non-linearity. Each neuron in this layer corresponds to a class $(k)$ and its activation is treated as the probability predicted by the model for that class.  The weights connecting the previous layer to this neuron $(\bm{w}_k)$ can be considered as the template of the class $k$ learned by the \textit{Trained} network. This is because the predicted class probability is proportional to the alignment of the pre-final layer's output with the template $(\bm{w}_k)$. The predicted probability peaks when the pre-final layer's output is a positive scaled version of this template $(\bm{w}_k)$. On the other hand, if the output of the pre-final layer is misaligned with the template $\bm{w}_k$, the confidence predicted for class $k$ is reduced. Therefore, we treat the weights $\bm{w}_k$ as the class template for class $k$ and compute the similarity between classes $i$ and $j$ as:
\begin{equation}
    C(i,j)=\frac{\bm{w}_i^T\bm{w}_j}{ \norm{\bm{w}_i} \norm{\bm{w}_j} }
    \label{eqn:class-sim-mat}
\end{equation}
Since the elements of the concentration parameter  have to be positive real numbers, we further perform a min-max normalization over each row of the class similarity matrix. The visualization of the class similarity matrix calculated from a CIFAR-$10$ trained model is shown in Figure~\ref{fig:similarity-matrix-cifar}.
%\vspace{-0.1in}
\subsection{Crafting \textit{Data Impressions} via Dirichlet Sampling}
\label{subsec:crafting}
Once the parameters $K$ and $\bm{\alpha}^k$ of the Dirichlet distribution are obtained for each class $k$, we can sample class probability (softmax) vectors, which respect the class similarities as learned by the \textit{Trained} network. Using the optimization procedure in eq.~(\ref{eqn:dir-ci}) we obtain the input representations corresponding to these sampled output class probabilities.
Let $Y^{k} = [\bm{y}_1^{k},\bm{y}_2^{k},\ldots,\bm{y}_N^{k}] \in \ \mathbb{R}^{K \times N}$, be the $N$ softmax vectors corresponding to class $k$, sampled from $Dir(K,\bm{\alpha}^k)$ distribution. Corresponding to each sampled softmax vector $\bm{y}_i^{k}$, we can craft a Data Impression ${\bar{x}_i}^k$, for which the \textit{Trained} network predicts a similar softmax output. We achieve this by optimizing the objective shown in eq.~(\ref{eqn:dir-ci}). We initialize $\bar{x}^k_i$ as a random noisy image and update it over multiple iterations till the cross-entropy loss between the sampled softmax vector $(\bm{y}_i^{k})$ and the softmax output predicted by the \textit{Trained} model $T$,  is minimized.
\begin{equation}
    \bar{x_i}^k=\underset{x}{\mathrm{argmin}} \ L_{CE}(\bm{y}_i^{k},T(x,\theta_T,\tau))
    \label{eqn:dir-ci}
\end{equation}
where $\tau$ is the temperature used in the softmax layer. The process is repeated for each of the $N$ sampled softmax probability vectors in $Y^{k}$, $k \in \{1 \ldots K\}$.
%%%%%%%%%%%%%%%%%%%%%%
\normalem
%%%%%%%%%%%% required to remove the underlines for the conditional statements in the algorithm (which is default when we use ulem package)
\begin{algorithm}[htp]
\SetAlgoLined
\SetKwInOut{Input}{Input}  
\Input{Trained classifier $T$ \\$N$: number of DIs crafted per category,\\ $[\beta_1,\beta_2,...,\beta_B]$: $B$ scaling factors,\\
$\tau$: Temperature 
}
\SetKwInOut{Output}{Output}  
\Output{$\bar{X}$: Data Impressions}
Obtain $K$: number of categories from T

Compute the class similarity matrix\\ $C=[\textbf{c}_1^T,\textbf{c}_2^T,\ldots,\textbf{c}_K^T]$ as in eq.~(\ref{eqn:class-sim-mat})

$\bar{X} \leftarrow \emptyset$%, $P \leftarrow \Phi$

\For{k=1:K }{

Set the concentration parameter $\bm{\alpha}^k =\textbf{c}_k$

    \For{b=1:B}{

        \For{n=1:$\left \lfloor{N/B}\right \rfloor $}{
        
            Sample ${\bm{y}_n^k} \sim Dir(K,\beta_b \times \bm{\alpha}^k)$

            Initialize $\bar{x}_n^k$ to random noise and craft $\bar{x}_n^k = \underset{x}{\mathrm{argmin}} \ L_{CE}({\bm{y}_n^k},T(x,\theta_T,\tau))$

            $\bar{X} \gets \bar{X} \cup \bar{x}_n^k$ 
            
        }

    }
    
}

\caption{Generation of Data Impressions}
\label{algo:zskd}
\end{algorithm}
%%%%%%%%%%
\ULforem
%%%% required to restore the default
\begin{figure*}[htp]
\centering
 \includegraphics[width=\textwidth, height=0.26\textwidth]{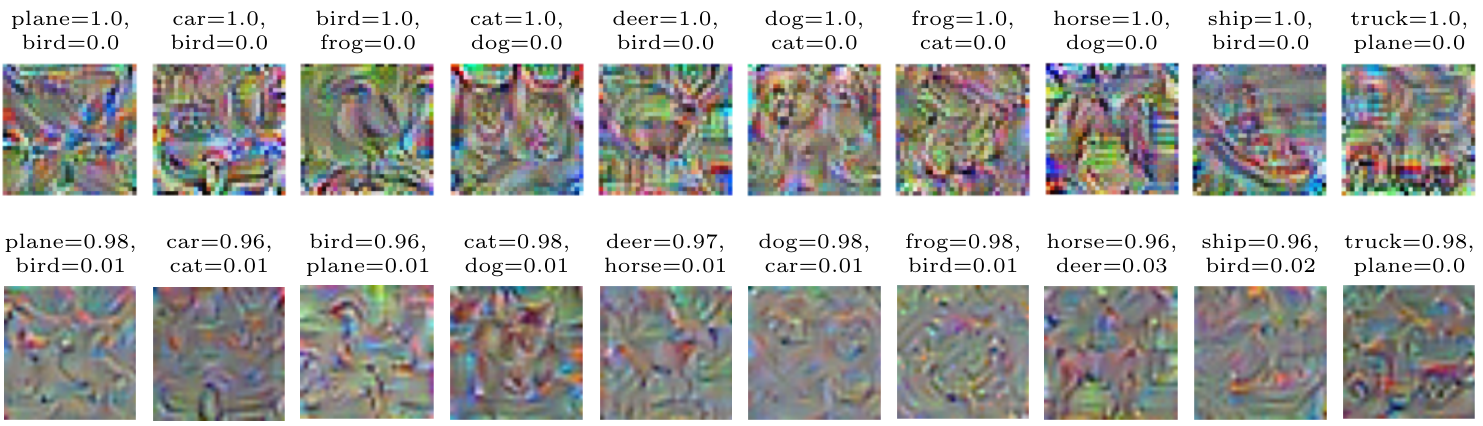} 
\caption{Visualizing the DIs synthesized from the \Te{} model trained on the CIFAR-$10$ dataset for different choices of output softmax vectors (i.e., output class probabilities). Note that the figure shows $2$ DIs per class in each column, each having a different spread over the labels. However, only the top-$2$ confidences in the sampled softmax corresponding to each $DI$ are mentioned on top for clarity.}
%Please note that there is no explicit objective for encouraging these pseudo samples to be visually closer to the actual training data samples, and yet some of the samples show striking patterns visually very similar to actual object shapes (e.g., bird, car, cat, dog, and deer in the first/second rows).}
\label{fig:vis-mnist}
\end{figure*}

\textbf{Scaling Factor ($\beta$)}: The probability density function of the Dirichlet distribution for $K$ random variables is a $K-1$ dimensional probability simplex that exists on a $K$ dimensional space. In addition to parameters $K$ and $\bm{\alpha}$ as discussed in section~\ref{subsec:dirichlet-modelling}, it is important to discuss the significance of the range of $\alpha_i  \in  \bm{\alpha}$ , in controlling the density of the distribution. When $\alpha_i < 1, \  \forall  i \ \epsilon[1,K]$, the density congregates at the edges of the simplex~\cite{balakrishnan2004primer,lin2016dirichlet}. As their values increase (when $\alpha_i > 1,  \forall i \in [1,K]$), the density becomes more concentrated on the center of the simplex~\cite{balakrishnan2004primer,lin2016dirichlet}. Thus, we define a scaling factor $(\bm{\beta})$ which can control the range of the individual elements of the concentration parameter, which in turn decides regions in the simplex from which sampling is performed. This becomes a hyper-parameter for the algorithm. Thus the actual sampling of the probability vectors happen from $p(\bm{s})= Dir( K, \beta \times \bm{\alpha})$. $\beta$ intuitively models the spread of the Dirichlet distribution and acts as a scaling parameter atop $\bm\alpha$ to yield the final concentration parameter (prior). $\beta$ controls the $l_1$-norm of the final concentration parameter which, in turn, is inversely related to the variance of the distribution. Variance of the sampled simplexes is high for smaller values of $\beta$ . However very low values for $\beta$  (e.g. $0.01$), in conjunction with the chosen $\bm\alpha$, result in highly sparse softmax vectors concentrated on the extreme corners of the simplex, which is equivalent to generating class impressions (see Fig.~\ref{fig:ci-vs-di}). As per the ablation studies, $\beta$ values of 0.1, 1.0 or a mix of these are in general favorable since they encourage higher diversity (variance) and at the same time does not result in highly sparse vectors. Our proposed approach for generating Data Impressions from a \textit{Trained} classifier is presented in Algorithm~\ref{algo:zskd}. 

Some of the resulting DIs are presented in Figure~\ref{fig:vis-mnist} %for MNIST,
for the CIFAR-$10$ dataset. Note that the figures show $2$ DIs per category. Also, note that the top-$2$ confidences in the sampled softmax corresponding to each DI are mentioned on top. We observe that the DIs are visually far away from the actual data samples of the dataset. However, some of the DIs synthesized from peaky softmax vectors (e.g. the bird, cat, car, and deer in the first row) contain clearly visible patterns of the corresponding objects. The observation of the DIs being visually far away from the actual data samples is understandable, since the objective to synthesize them (eq.~(\ref{eqn:dir-ci})) pays no explicit attention to visual detail. %However, it is interesting to 
%%%%%%%%%%%%%%%%%%%%%%%%
%\vspace{-0.1in}

\section{Applications of Data Impressions and Experimental Evaluation}
\label{sec:applications}
The generated Data Impressions through the proposed approach can be utilized for several applications in the absence of training data, which we discuss in detail. We specifically study the application of Data Impression for multiple important CV/ML tasks, viz., Zero-shot knowledge distillation, Unsupervised Domain Adaptation, Continual Learning and Data-free UAP Generation. Here, for each application area, we first introduce the problem and describe how the extracted data-impressions can be utilized towards these tasks. Subsequently, we provide a detailed experimental evaluation to justify the utility of DIs in the given task.
\subsection{Zero-Shot Knowledge Distillation}
\label{subsec:kd}
Transferring the generalization ability of a large, complex \Te{} $(T)$ deep neural network to a relatively simpler \St{} $(S)$ network can be achieved using the class probabilities produced by a \Te{} as ``soft targets"~\cite{hinton2015distilling} for training the \St{}. For this transfer, most of the existing approaches require access to the original training data consisting of tuples of input data and targets $(x,y) \in \mathbb{D}$. Let $T$ be the \Te{} network with learned parameters $\theta_T $ and $S$ be the \St{} with parameters $\theta_S$, note that in general $\lvert \theta_S \rvert \ll \lvert \theta_T \rvert$. Knowledge distillation methods train the \St{} via minimizing the following objective $(L)$ with respect to the parameters $\theta_S$ over the training samples ${(x,y) \in \mathbb{D}}$
\begin{equation}
    L = \sum_{ (x,y) \in \mathbb{D} }L_{KD}(S(x,\theta_S,\tau),T(x,\theta_T,\tau))  +  \lambda  L_{CE}({\hat{y}}_S, y)
    \label{eqn:kd}
    %\underset{(x,y) \in \mathbb{X} \times \mathbb{Y} }{\text{minimize}}
\end{equation}
 $L_{CE}$ is the cross-entropy loss computed on the labels ${\hat{y}}_S$ predicted by the \St{} and their corresponding ground truth labels $y$. $L_{KD}$ is the distillation loss (e.g. cross-entropy or mean square error) comparing the soft labels (softmax outputs) predicted by the \St{} against the soft labels predicted by the \Te{}. $T(x,\theta_T)$ represents the softmax output of the \Te{} and $S(x,\theta_S)$ denotes the softmax output of the \St{}. Note that, unless it is mentioned, we use a softmax temperature of $1$. If we use a temperature value $(\tau)$ different from $1$, we represent it as $S(x,\theta_S,\tau)$ and $T(x,\theta_T,\tau)$ for the remainder of the paper. $\lambda$ is the hyper-parameter to balance the two objectives.

Once we craft the Data Impressions (DI) $(\bar{X})$ from the \Te{} model using Algorithm~\ref{algo:zskd}, we treat them as the `Transfer set' and perform the knowledge distillation. Note that we use only the distillation loss $L_{KD}$ as shown in eq.~(\ref{eqn:zskd}). We ignore the cross-entropy loss from the general Distillation objective (eq.~(\ref{eqn:kd})) since there is only minor to no improvement in the performance and it reduces the burden of hyper-parameter $\lambda$. 
\begin{align}
    \theta_S=\underset{\theta_S}{\mathrm{\:argmin\:}} \sum_{\bar{x} \in \bar{X}} L_{KD}(T(\bar{x},\theta_T,\tau), S(\bar{x},\theta_S,\tau))
    \label{eqn:zskd}
\end{align}
Thus we generate a diverse set of pseudo training examples that can provide with enough information to train the \St{} model via Dirichlet sampling. In the subsequent sections, we discuss the experimental evaluation of the proposed data-free knowledge transfer framework over a set of benchmark object recognition datasets

%\subsubsection{Zero-Shot Knowledge Distillation}
%\label{subsec:zskd}

\subsubsection{Experimental Setup and Datasets}
We experimentally evaluate our proposed Zero-Shot Knowledge Distillation (ZSKD) approach on MNIST \cite{lecun1998gradient}, Fashion MNIST (FMNIST) \cite{xiao2017fashion}, and CIFAR-$10$ \cite{krizhevsky2009learning}. Here, we provide detailed experimental setup for each of these three datasets.

%\paragraph{\textbf{MNIST}}
%{\color{blue}
\textbf{MNIST:}
\label{subsec:mnist}
%The MNIST dataset has $60000$ training images and $10000$ test images of handwritten digits. 
We consider Lenet-$5$ for the \Te{} model and Lenet-$5$-Half for \St{} model similar to \cite{dfkd-nips-lld-17}. The Lenet-$5$ Model contains $2$ convolution layers and pooling which is followed by three fully connected layers.  Lenet-$5$ is modified to make Lenet-$5$-Half by taking half the number of filters in each of the convolutional layers. The \Te{} and \St{} models have $61706$ and $35820$ parameters respectively. Input images are resized from $28 \times 28$ to $32 \times 32$ and the pixel values are normalized to be in $[0, 1]$ before feeding into the models.

%\paragraph{\textbf{Fashion-MNIST}}
%{\color{blue}
\textbf{Fashion-MNIST:}
\label{subsec:fmnist}
%In comparison to MNIST, this dataset is more challenging and contains images of fashion products. %The training and testing set has $60000$ and $10000$ images respectively. 
Similar to MNIST, we consider Lenet-$5$ and Lenet-$5$-Half as \Te{} and \St{} model respectively where each input image is resized from dimension $28 \times 28$ to $32 \times 32$. 

%\paragraph{\textbf{CIFAR-10}}
%{\color{blue}
\textbf{CIFAR-10:}
\label{subsec:cifar}
Unlike MNIST and Fashion MNIST, this dataset contains RGB images of dimension $32 \times 32 \times 3$. %The dataset contains $60000$ images from $10$ classes, where each class has $6000$ images. Among them, $50000$ images are form the training set and rest of the $10000$ images compose the test set. 
We take AlexNet~\cite{krizhevsky2012imagenet} as \Te{} model which is relatively large in comparison to LeNet-$5$. Since the standard AlexNet model is designed to process input of dimension $227 \times 227 \times 3$, we need to resize the input image to this large dimension. To avoid that, we have modified the standard AlexNet to accept $32 \times 32 \times 3$ input images. The modified AlexNet contains $5$ convolution layers with BatchNorm~\cite{batchnorm-icml-2015} regularization. Pooling is also applied on convolution layers $1, 2,$ and $5$. The deepest three layers are fully connected. AlexNet-Half is derived from this AlexNet by taking half of convolutional filters and half of the neurons in the fully connected layers except in the classification layer which has number of neurons equal to number of classes. The AlexNet-Half architecture is used as the \St{} model. The \Te{} and \St{} models have $1.65 \times 10^6$ and $7.23 \times 10^5$ parameters respectively.

\subsubsection{Implementation Details}
As all the experiments in these three datasets are dealing with classification problems with $10$ categories each, value of the parameter $K$ in all our experiments is $10$. For each dataset, we first train the \Te{} model over the available training data using the cross-entropy loss. Then we extract a set of Data Impressions $(DI)$ from it via modelling its softmax output space as explained in sections~\ref{subsec:dirichlet-modelling} and \ref{subsec:crafting}. Finally, we choose a (light weight) \St{} model and train over the transfer set (DI) using eq.~(\ref{eqn:zskd}). 

We consider two $(B=2)$ scaling factors, $\beta_{1} = 1.0$ and $\beta_{2} = 0.1$ across all the datasets, i.e., for each dataset, half the Data Impressions are generated with $\beta_1$ and the other with $\beta_2$. However we observed that one can get a fairly decent performance with a choice of beta equal to either $0.1$ or $1$ (even without using the mixture of Dirichlet) across the datasets. A temperature value $(\tau)$ of $20$ is used across all the datasets. Also, since the proposed approach aims to achieve better generalization, it is a natural choice to augment the crafted Data Impressions while performing the distillation. We augment the samples using regular operations such as scaling, translation, rotation, flipping etc. which has proven useful in further boosting the model performance \cite{dao2018kernel}. 

In section ~\ref{subsec:zskd-results}, we show the ZSKD results on the three benchmark datasets. In the subsequent sections, we investigate in detail, the effect of transfer set size, i.e., the number of Data Impressions on the performance of the \St{} model (sec.~\ref{subsec:effect-size}), compare the ZSKD results when used with Class Impressions \cite{mopuri2018ask} (sec.~\ref{subsec:class-data}), apply ZSKD on large architectures (sec.~\ref{subsec:zskd_larger_arch}) and finally show that DIs preserve adversarial robustness in the ZSKD framework (sec.~\ref{subsec:di-robustness}).

\subsubsection{Results and Discussion}
\label{subsec:zskd-results}
The performance of Zero-Shot Knowledge Distillation for the MNIST, Fashion-MNIST, and CIFAR-10 datasets is presented in Tables~\ref{tab:mnist},~\ref{tab:fmnist}, and~\ref{tab:cifar} respectively. Note, that in order to understand the effectiveness of the proposed ZSKD, the tables also show the performance of the \Te{} and \St{} models trained over actual data samples along with a comparison against existing distillation approaches. Teacher-CE denotes the classification accuracy of the \Te{} model trained using the cross-entropy (CE) loss, Student-CE denotes the performance of the \St{} model trained with all the training samples and their ground truth labels using cross-entropy loss. Student-KD denotes the accuracy of the \St{} model trained using the actual training samples through Knowledge Distillation (KD) from \Te{}. Note that this result may act as an upper bound for the data-free distillation approaches.
\begin{table}[htp]
\caption{Performance of the proposed ZSKD framework on the MNIST dataset.}
\label{tab:mnist}
\centering
\begin{tabular}{|c|c|}
\hline
\multicolumn{1}{|c|}{\textbf{Model}} & \multicolumn{1}{c|}{\textbf{Performance}} \\ \hline
Teacher-CE                             &       99.34                                    \\ \hline
Student-CE                           &       98.92                                      \\ \hline
\makecell{Student-KD \cite{hinton2015distilling}\\ 60K original data}           &       99.25                                    \\ \hline
\makecell{\cite{kimura2018few} \\200 original data}                           &   86.70                                        \\ \hline
\makecell{\cite{dfkd-nips-lld-17}\\ (uses meta data)}                           &    92.47                                       \\ \hline
\makecell{\textbf{ZSKD} (Ours) \\($24000$ DIs, and no original data)}                         &       98.77                                    \\ \hline
%extra                                & extra                                     \\ \hline
\end{tabular}
\end{table}

Table~\ref{tab:mnist} presents our results on MNIST, and compares them with existing approaches. It is clear that the proposed Zero-Shot Knowledge Distillation (ZSKD) outperforms the existing few data \cite{kimura2018few} and data-free counterparts \cite{dfkd-nips-lld-17} by a great margin. Also, it performs close to the full data (classical) Knowledge Distillation while using only $24000$ \textit{DI}s, i.e., $40\%$ of the the original training set size.

\begin{table}[htp]
\caption{Performance of the proposed ZSKD framework on the Fashion MNIST dataset.}
\label{tab:fmnist}
\centering
\begin{tabular}{|c|c|}
\hline
\multicolumn{1}{|c|}{\textbf{Model}} & \multicolumn{1}{c|}{\textbf{Performance}} \\ \hline
Teacher-CE                              &     90.84                                      \\ \hline
Student-CE                           &     89.43                                     \\ \hline
\makecell{Student-KD \cite{hinton2015distilling}\\ 60K original data}          &     89.66                                      \\ \hline
\makecell{\cite{kimura2018few} \\200 original data}                        &                 72.50                           \\ \hline
%Existing-2                           &                                           \\ \hline
\makecell{\textbf{ZSKD} (Ours) \\($48000$ DIs, and no original data)}                          &     79.62                                     \\ \hline
%extra                                & extra                                     \\ \hline
\end{tabular}
\end{table}

Table~\ref{tab:fmnist} presents our results for Fashion-MNIST and compares them with the existing approaches. Similar to MNIST, ZSKD outperforms the existing few data knowledge distillation approach~\cite{kimura2018few} by a large margin, and performs close to the classical knowledge distillation scenario~\cite{hinton2015distilling} with all the training samples.

 %For architectural details of the teacher and the student nets, please refer to the supplementary document.

\begin{table}[htp]
\caption{Performance of the proposed ZSKD framework on the CIFAR-$10$ dataset.}
\centering
\label{tab:cifar}
\begin{tabular}{|c|c|}
\hline
\multicolumn{1}{|c|}{\textbf{Model}} & \multicolumn{1}{c|}{\textbf{Performance}} \\ \hline
Teacher-CE                              &     83.03                                      \\ \hline
Student-CE                           &     80.04                                      \\ \hline
\makecell{Student-KD \cite{hinton2015distilling}\\ 50K original data}          &     80.08                                      \\ \hline
%Student KD (Hinton et al.)           &     79.98                                      \\ \hline
%Existing-1                           &                                           \\ \hline
%Exiating-2                           &                                           \\ \hline
%ZSKD (Ours)                          &     69.56                                      \\ \hline
%extra                                & extra                                     \\ \hline
\makecell{\textbf{ZSKD} (Ours) \\($40000$ DIs, and no original data)}                          &     69.56                                     \\ \hline
\end{tabular}
\end{table}
\begin{figure*}[htp]
\centering
\noindent\begin{minipage}{\textwidth}
  \centering
  \includegraphics[width=.31\textwidth]{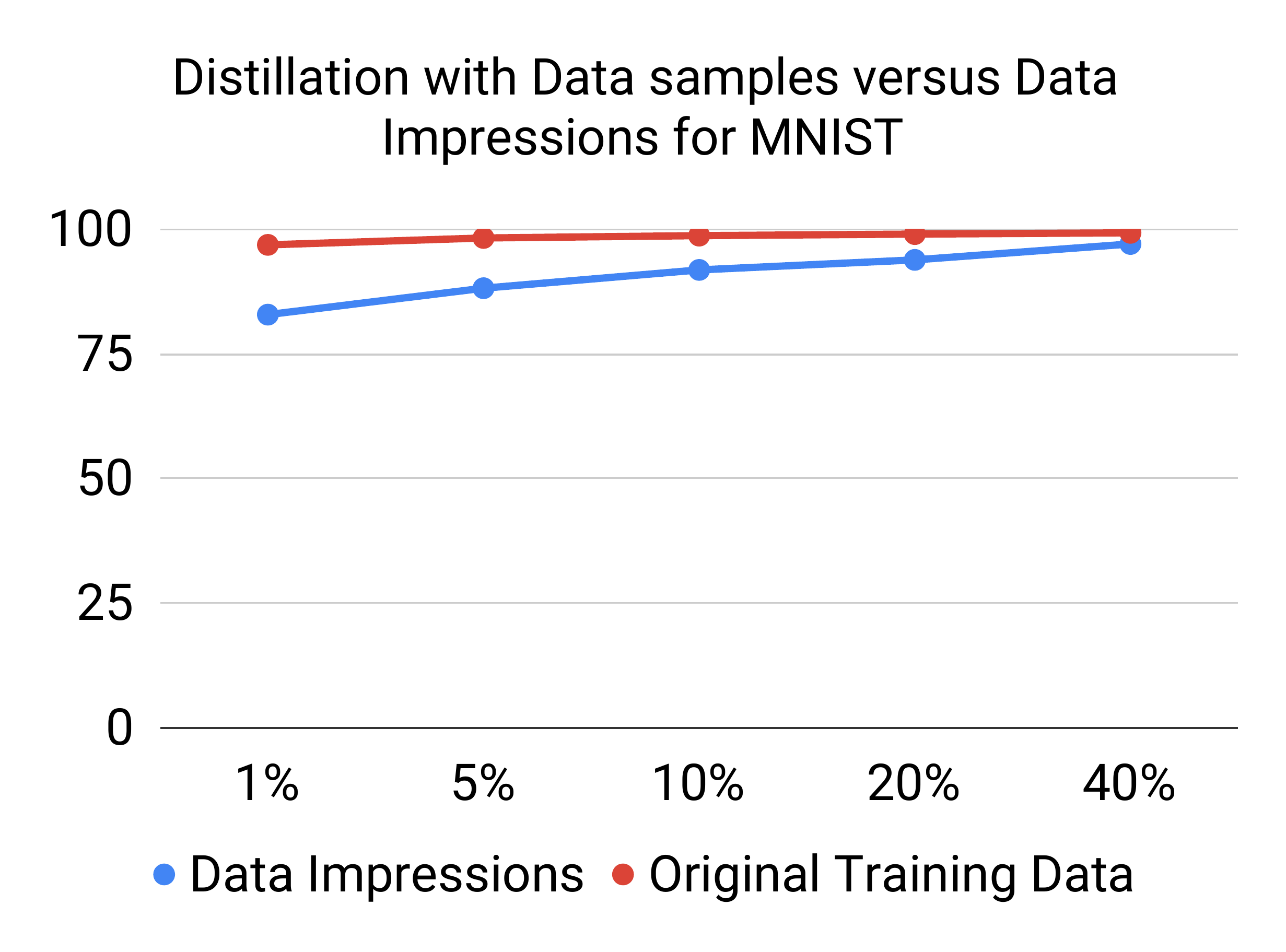}\hfill
  \includegraphics[width=.31\textwidth]{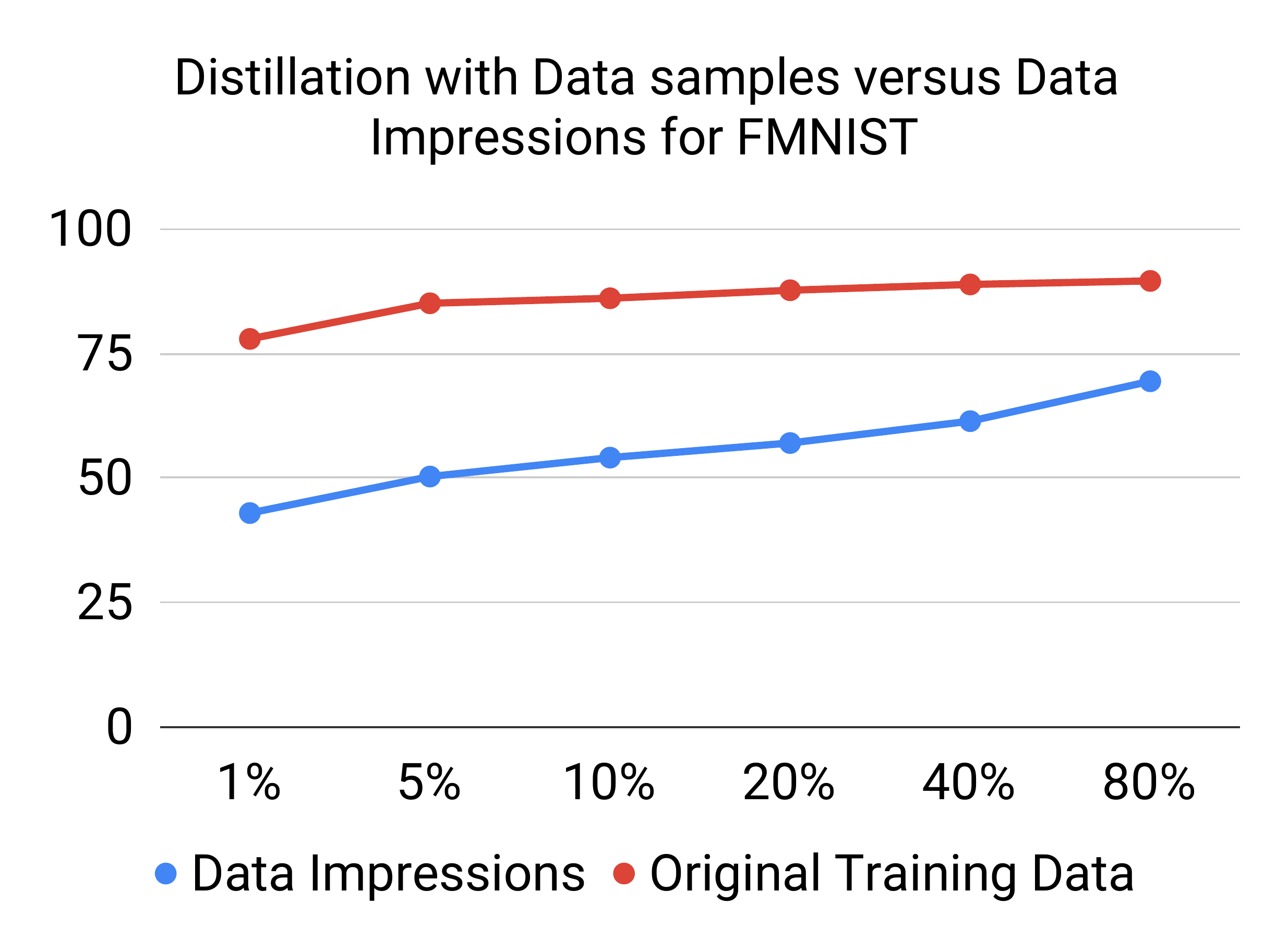}\hfill
  \includegraphics[width=.31\textwidth]{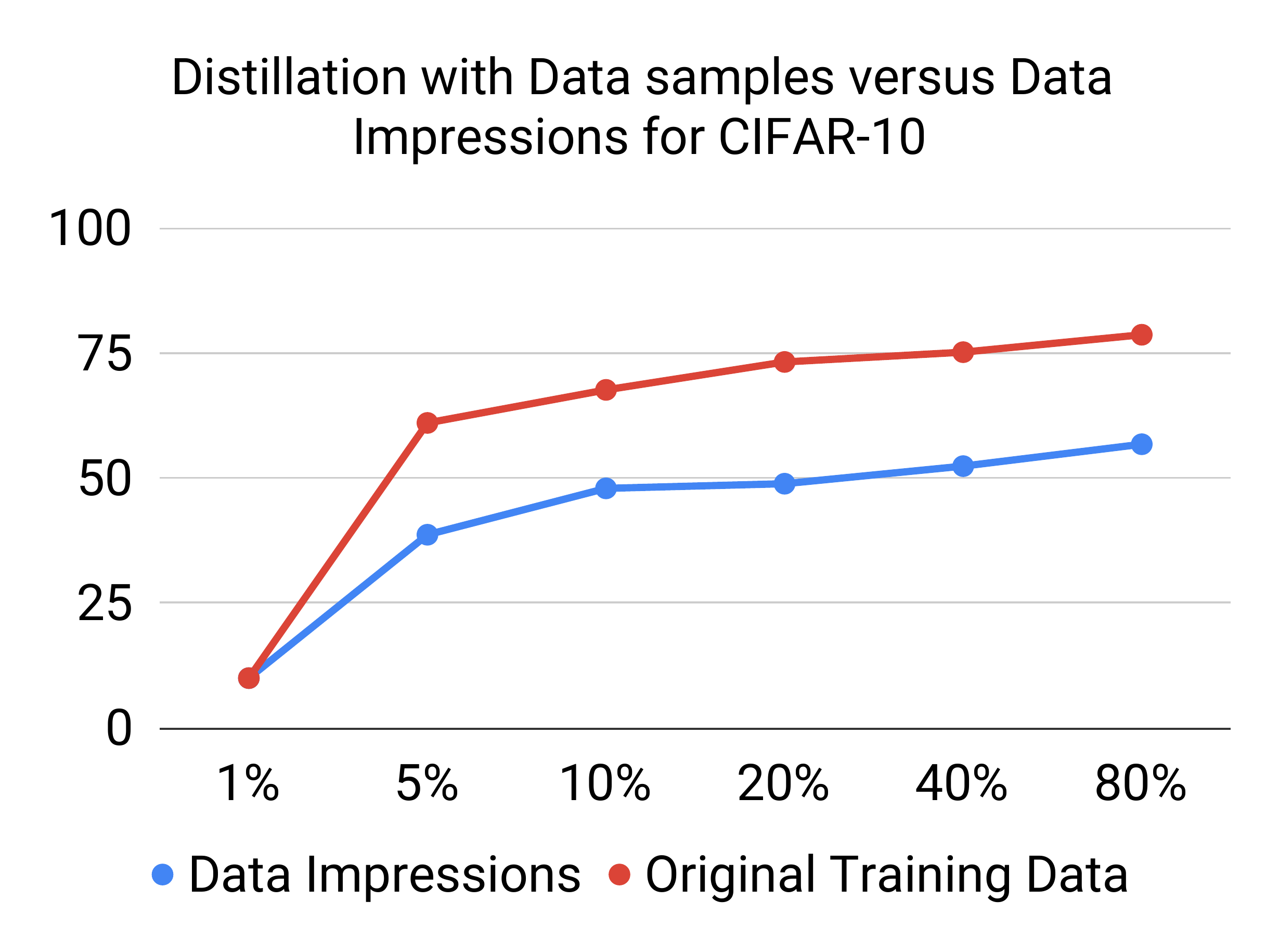}\hfill
  \vspace{0.01\textwidth}
\end{minipage}
\caption{Performance (Test Accuracy) comparison of Data samples versus Data Impressions (without augmentation). Note that the x-axis denotes the number of DIs or original training samples (in $\%$) used for performing Knowledge Distillation with respect to the size of the training data.% note that the results are without the data augmentation.
}
\label{fig:data-vs-di}
\end{figure*}
\begin{figure*}[htp]
\centering
\noindent\begin{minipage}{\textwidth}
  \centering
  \includegraphics[width=.32\textwidth]{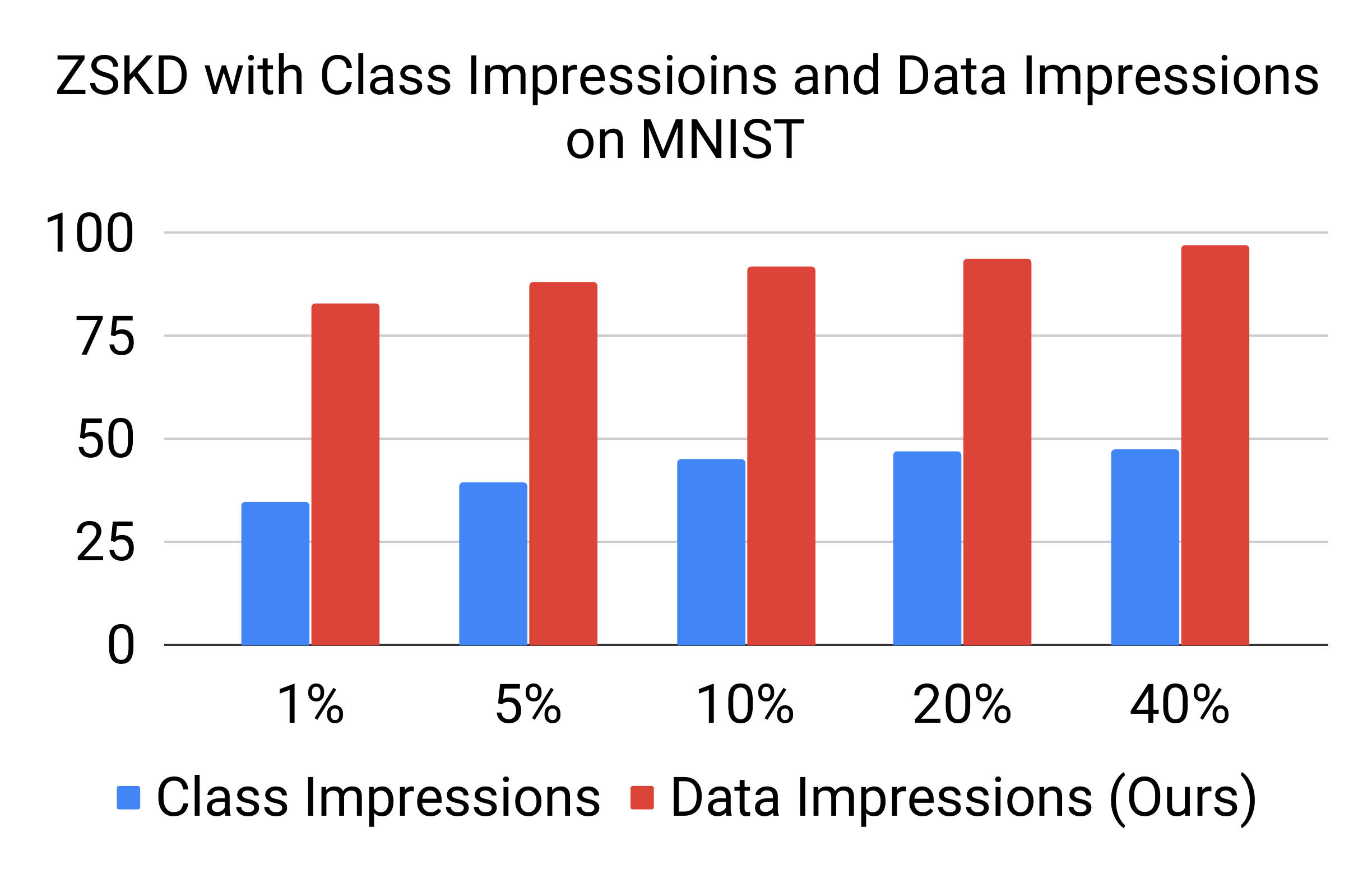}\hfill
  \includegraphics[width=.32\textwidth]{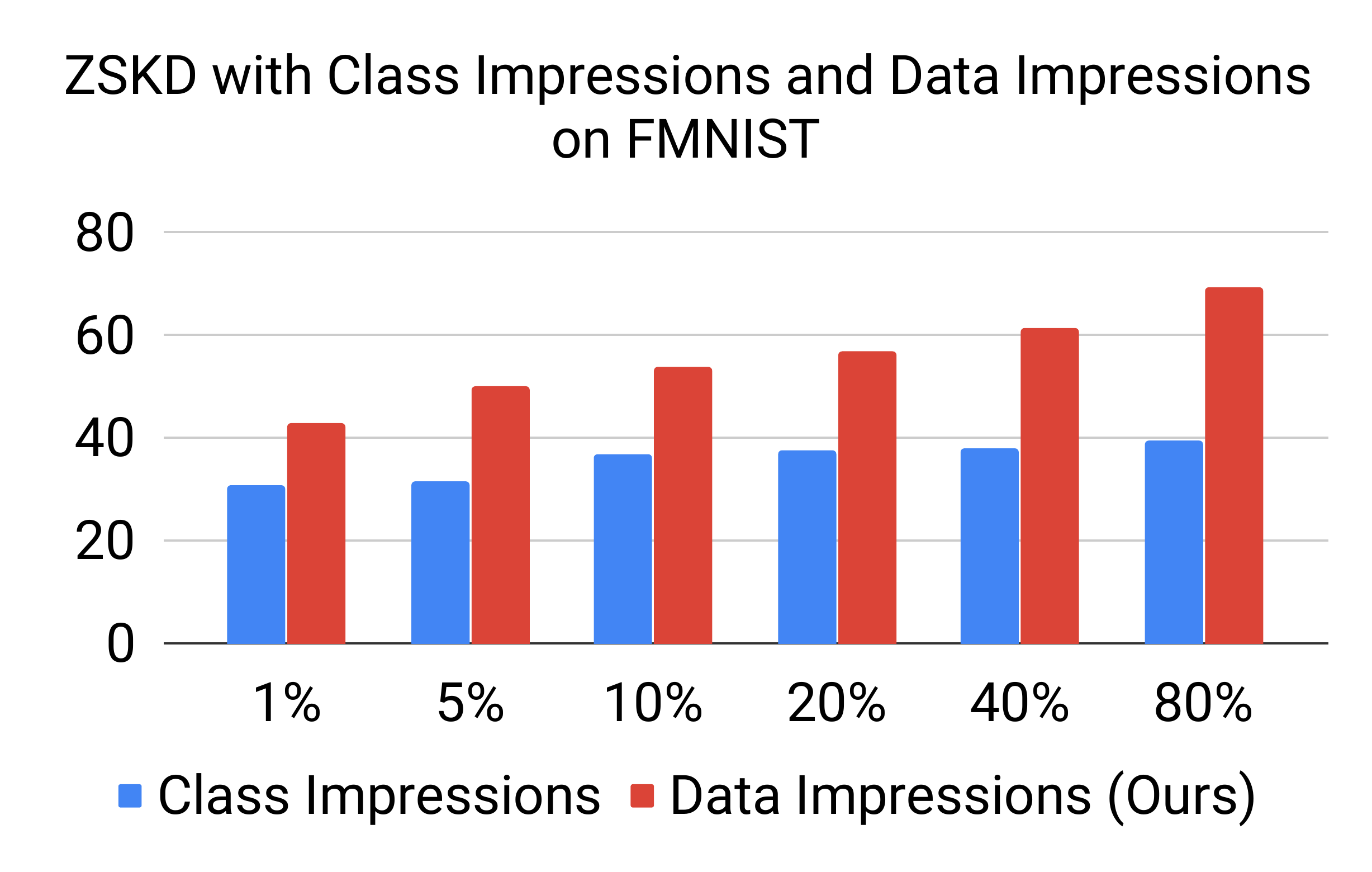}\hfill
  \includegraphics[width=.32\textwidth]{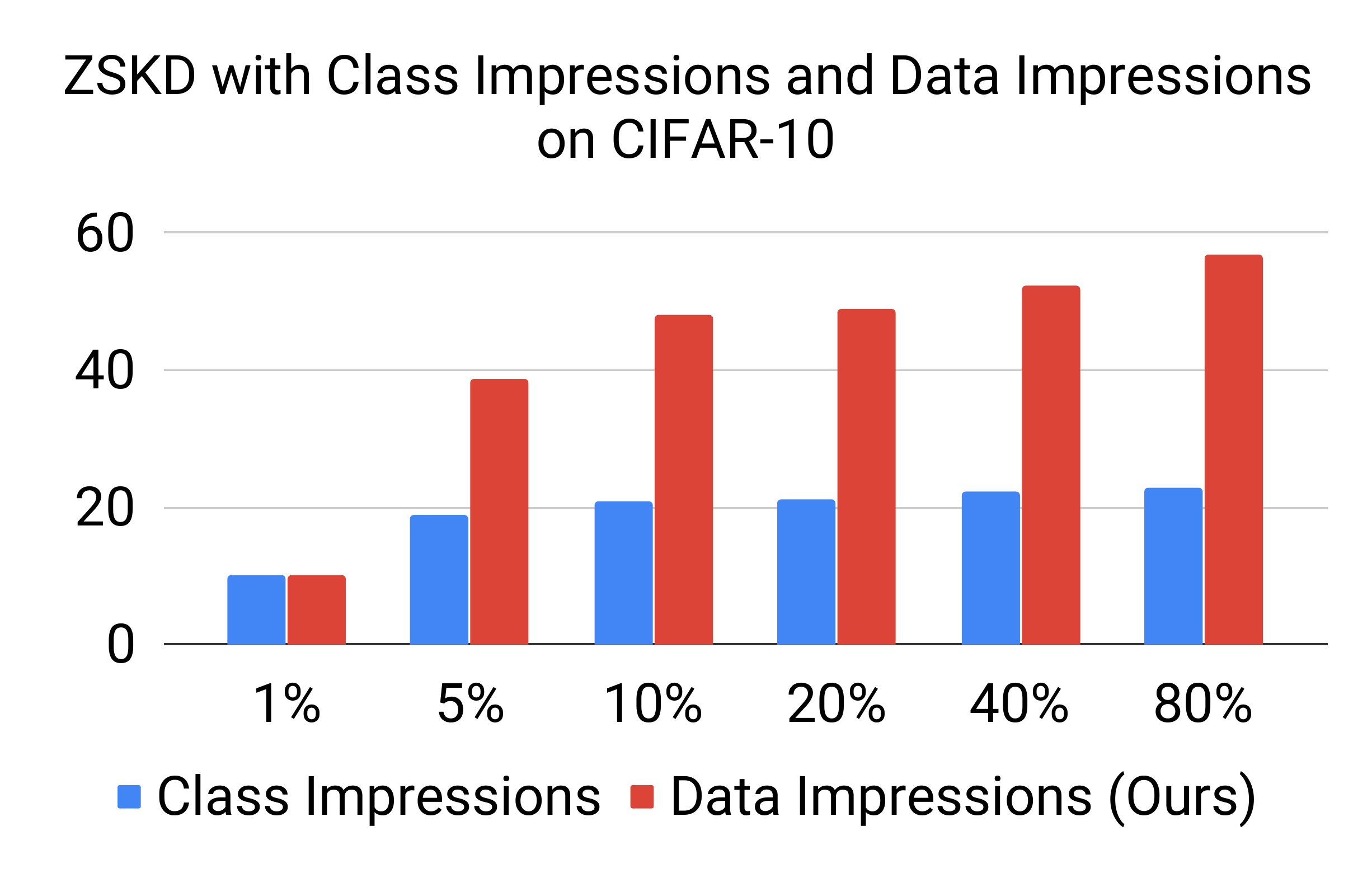}\hfill
  \vspace{0.01\textwidth}
\end{minipage}
\caption{Performance (Test Accuracy) comparison of the ZSKD with Class Impressions~\cite{mopuri2018ask} and proposed Data Impressions (without augmentation). %The panels show the test accuracy of the distilled \St{} model over multiple datasets. 
Note that the x-axis denotes the number of DIs or CIs (in \%) used for performing Knowledge Distillation with respect to the training data size. 
}
\label{fig:ci-vs-di} 
\end{figure*}
Table~\ref{tab:cifar} presents the results on CIFAR-$10$ dataset. It can be observed that the proposed ZSKD approach can achieve knowledge distillation with the Data Impressions that results in performance competitive to that realized using the actual data samples. Since the underlying target dataset is relatively more complex, we use a bigger transfer set containing $40000$ DIs. However, the size of this transfer set containing DIs is still $20\%$ smaller than that of the original training set size used for the classical knowledge distillation \cite{hinton2015distilling}.

\subsubsection{Effect of Transfer Set Size}
\label{subsec:effect-size}
In this subsection, we investigate the effect of transfer set size on the performance of the distilled \St{} model. We perform the distillation with different number of Data Impressions such as $\{1\%,\: 5\%,\: 10\%,\: \ldots, 80\%\}$ of the training set size. Figure~\ref{fig:data-vs-di} shows the performance of the resulting \St{} model on the test set for all the datasets. For comparison, the plots present performance of the models distilled with the equal number of actual training samples from the dataset. It is observed that, as one can expect, the performance increases with size of the transfer set. %Interestingly, even a small number of Data Impressions (e.g. $20\%$ of the training set size) are sufficient to provide a competitive performance, though the improvement in performance gets quickly saturated. 
Also, note that the initial performance (with smaller transfer set) reflects the complexity of the task (dataset). For simpler datasets such as MNIST, smaller transfer sets are sufficient to achieve competitive performance. In other words, small number of Data Impressions can do the job of representing the patterns in the dataset. As the dataset becomes complex, more number of Data Impressions need to be generated to capture the underlying patterns in the dataset. Note that similar trends are observed in the distillation with the actual training samples as well.

\begin{table*}[htp]
\caption{Performance measures to evaluate the robustness transfered under distillation using Data Impressions for different datasets. $A_{nat}$ denotes the accuracy obtained on unperturbed data whereas $A_{adv}$ denotes adversarial accuracy i.e. the performance of the model on the perturbed data. F.R. is the `fooling rate' which describes the amount of samples whose labels got changed after adversarial attack. All the numbers shown are in \%.}
\label{tab:robustness-mnist}
\setlength{\tabcolsep}{8pt} % Default value: 6pt
\renewcommand{\arraystretch}{1.1} % Default value: 1
\centering
\begin{tabular}{|c|l|c|c|c|c|c|c|c|c}
\cline{1-9}
\multirow{3}{*}{\textbf{Dataset}} & \multirow{3}{*}{\textbf{Model}} & \multicolumn{7}{c|}{\textbf{Performance Measures (in \%)}} &  \\ \cline{3-9}
 &  & Natural & \multicolumn{2}{c|}{FGSM} & \multicolumn{2}{c|}{IFGSM} & \multicolumn{2}{c|}{PGD} &  \\ \cline{3-9}
 &  & A$_{nat}$ \(\uparrow\) & A$_{adv}$ \(\uparrow\)& F.R. \(\downarrow\)& A$_{adv}$ \(\uparrow\)& F.R. \(\downarrow\)& A$_{adv}$ \(\uparrow\)& F.R. \(\downarrow\)&  \\ \cline{1-9}
\multirow{4}{*}{MNIST} & Non-robust teacher & 99.34 & 21.01 & 79.45 & 1.65 & 98.87 & 0.51 & 99.97 &  \\ \cline{2-2}
 & Student using DIs from non-robust teacher & 98.77 & 37.4 & 63.44 & 8.52 & 92.39 & 3.06 & 97.86 &  \\ \cline{2-9}
 & Robust teacher & 98.01 & 97.06 & 2.24 & 95.04 & 5.66 & 95.0 & 5.6 &  \\ \cline{2-2}
 & Student using DIs from robust teacher & 95.85 & 86.1 & 13.4 & 74.01 & 28.44 & 54.38 & 48.27 &  \\ \cline{1-9}
\multirow{4}{*}{Fashion-MNIST} & Non-robust teacher & 90.84 & 13.99 & 91.47 & 6.02 & 99.99 & 4.85 & 100.0 &  \\ \cline{2-2}
 & Student using DIs from non-robust teacher & 79.62 & 13.94 & 96.26 & 12.03 & 100.0 & 9.64 & 100.0 &  \\ \cline{2-9}
 & Robust teacher & 75.15 & 74.28 & 7.77 & 72.92 & 13.5 & 73.41 & 12.83 &  \\ \cline{2-2}
 & Student using DIs from robust teacher & 68.44 & 60.25 & 31.80 & 47.90 & 58.78 & 39.81 & 70.14 &  \\ \cline{1-9}
\multirow{4}{*}{CIFAR-10} & Non-robust teacher & 83.03 & 15.89 & 93.46 & 10.16 & 99.31 & 9.62 & 99.87 &  \\ \cline{2-2}
 & Student using DIs from non-robust teacher & 69.56 & 17.74 & 97.52 & 15.67 & 99.74 & 15.24 & 99.86 &  \\ \cline{2-9}
 & Robust teacher & 66.99 & 51.72 & 53.24 & 50.56 & 55.23 & 46.53 & 60.86 &  \\ \cline{2-2}
 & Student using DIs from robust teacher & 54.38 & 37.44 & 70.89 & 32.45 & 77.99 & 24.92 & 86.62 &  \\ \cline{1-9}
\end{tabular}
\end{table*}

\subsubsection{Class Versus Data Impressions}
\label{subsec:class-data}
Feature visualization works such as~\cite{backprop-iclrw-2014,guidedbackprop-iclrw-2015,olah2017feature,deep-dream-2015} attempt to understand the patterns learned by the deep neural networks in order to recognize the objects. These works reconstruct a chosen neural activation in the input space as one way to explain away the model's inference.

As described earlier, one of the recent works by \cite{mopuri2018ask} reconstructs samples of a given class for a downstream task of adversarial fooling. A random noise is optimized in the input space till it results in a one-hot vector (softmax) output. This means, their optimization to craft the representative samples would expect a one-hot vector in the output space. Hence, they call the reconstructions Class Impressions. Our reconstruction (eq.~(\ref{eqn:dir-ci})) is inspired from this, though we model the output space utilizing the class similarities perceived by the \Te{} model. Because of this, we argue that our modelling is closer to the original distribution and results in better patterns in the reconstructions, calling them Data Impressions of the \Te{} model. %Note that these impressions need not be closer in visual 

We compare these two varieties of reconstructions for the application of distillation. Figure~\ref{fig:ci-vs-di} demonstrates the effectiveness of Class and Data Impressions over three datasets. It is observed that the proposed Dirichlet modelling of the output space and the reconstructed impressions consistently outperform their class counterparts by a large margin. Also, in case of Class Impressions, the increment in the performance due to increased transfer set size is relatively small compared to that of Data Impressions. Note that for better understanding, the results are shown without any data augmentation while conducting the distillation. 

\subsubsection{Performance of ZSKD on Large Architectures}
\label{subsec:zskd_larger_arch}
In this section, we investigate the performance of ZSKD on popular network architectures, in addition to those studied in Sec.~\ref{subsec:zskd-results}. Note that, these architectures are also of much larger capacity than that of the models discussed earlier. % with relatively large architecture in comparison to the ones considered in our conference version. %that are taken for experiments in the previous sections. %In this section, we investigate the performance of ZSKD on popular large architectures, such as VGG and ResNet, for the CIFAR-10 dataset. 
Specifically, we perform experiments on VGG and Resnet architectures on the CIFAR-$10$ dataset.

\begin{table}[htp]
\centering
\caption{ZSKD performance using data impressions from VGG \Te{} architecture on CIFAR-$10$}
\begin{tabular}{|l|c|c|}
\hline
\textbf{Model}           & \textbf{Data-free} & \textbf{Performance (\%)} \\ \hline
VGG-19 (T)               & \xmark                  & 87.99                \\ \hline \hline
VGG-11 (S)- CE           & \xmark                  & 84.19                \\ \hline
VGG-11 (S)- KD~\cite{hinton2015distilling}           & \xmark                  & 84.93                \\ \hline
VGG-11 (S)- ZSKD (\textbf{Ours})    & \cmark                  & 74.10                \\ \hline \hline
Resnet-18 (S)- CE        & \xmark                  & 84.45                \\ \hline
Resnet-18 (S)- KD~\cite{hinton2015distilling}         & \xmark                  & 86.58                \\ \hline
Resnet-18 (S)- ZSKD (\textbf{Ours}) & \cmark                  & 74.76                \\ \hline
\end{tabular}

\label{vgg_teacher_zskd}
\end{table}

As shown in Table~\ref{vgg_teacher_zskd}, VGG-$19$ is taken as the Teacher network which is trained for $500$ epochs with a learning rate (lr) of $0.001$ and batch size of $512$. The knowledge from the trained Teacher is distilled into two different student models i.e. VGG-$11$ and Resnet-$18$. Their performance on original training data without (CE) and with distillation (KD) are also reported (the latter can be assumed as an upper bound). The data impressions are generated using adam optimizer with a batch size of $32$ and initial learning rate of $10$ with a $\beta$ mixture  of \{$0.1$, $1.0$\}. The learning rate is subsequently reduced linearly over the $1500$ iterations of optimization. The ZSKD performance on VGG-$11$ and Resnet-$18$ while distilling from the VGG-$19$ teacher with a learning rate of $0.001$ are $74.10\%$ and $74.76\%$ respectively.

\begin{table}[htp]
\centering
\caption{ZSKD performance using Data Impressions from Resnet-18 \Te{} architecture for CIFAR-$10$}
\begin{tabular}{|l|c|c|}
\hline
\textbf{Model}                & \textbf{Data-free} & \textbf{Performance (\%)} \\ \hline
Resnet-18 (T)                 & \xmark                  & 86.54                \\ \hline
Resnet-18-half (S)- CE        & \xmark                  & 85.51                \\ \hline \hline
Resnet-18-half (S)- KD~\cite{hinton2015distilling}        & \xmark                  & 86.31                \\ \hline
Resnet-18-half (S)- ZSKD (\textbf{Ours}) & \cmark                  & 81.10                \\ \hline
\end{tabular}

\label{tab:resnet-18-zskd}
\end{table}

We also perform the experiments with a different Teacher network architecture i.e. Resnet-$18$ which is trained with lr $0.01$, batch size of $512$ for $500$ epochs and obtain an accuracy of $86.54\%$. Here, we use Resnet-$18$-half as a student network which is formed by taking half the number of filters at each layer of Resnet-$18$. Similar to the previous experiment, we also report results with and without distillation using original training data as shown in Table~\ref{tab:resnet-18-zskd}. The data impressions are synthesized with a lr of $0.001$. Our ZSKD method obtains an accuracy of $81.10\%$ which is only $\approx 5\%$ less than the performance using the entire original training data (KD).
\begin{figure*}[htp]
\centering
 \includegraphics[width=\textwidth, ]{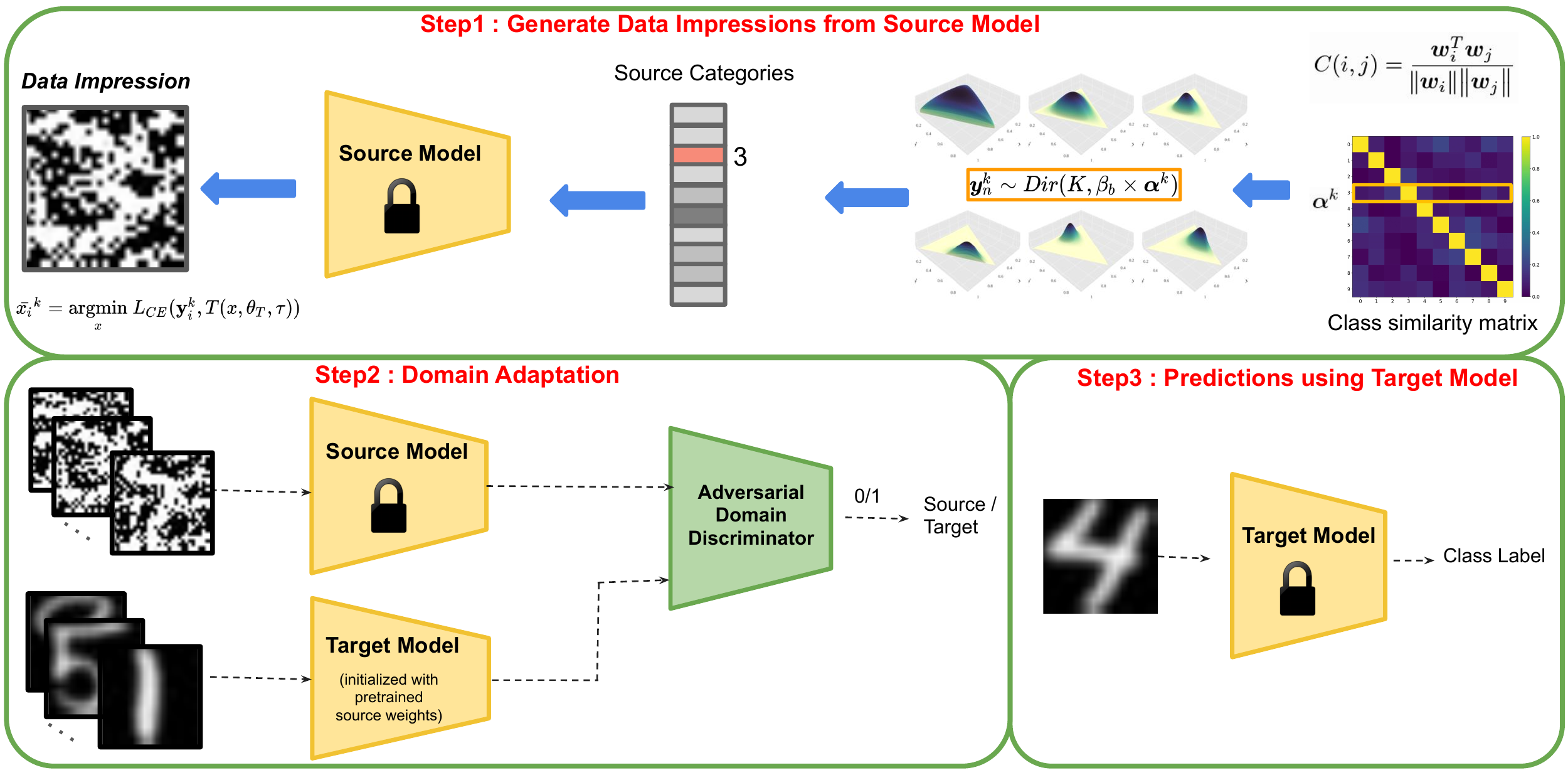}
\caption{Proposed Approach for Source Free Unsupervised Domain Adaptation using Data Impressions.}
\label{source-free-adaptation-approach}
\end{figure*}
\subsubsection{Investigating Adversarial Robustness of DI-Distilled models}
\label{subsec:di-robustness}

In this subsection, we demonstrate that Data Impressions are indeed close-to-true approximations of the training data by experimentally verifying that they capture the adversarial robustness property of an adversarially trained Teacher, and preserve it under zero-shot knowledge distillation.

While multiple works such as \cite{micaelli2019zero, chen2019data} have studied data-free approaches for training deep neural networks, to the best of our knowledge, we are the first to study the robustness properties of the resulting Student models. We empirically analysed the behaviour of Students that are distilled from normally trained versus adversarially trained Teachers. The distribution of adversarial samples (generated by perturbing natural images) would likely be different from the natural training data distribution. Therefore, it is critical to study if Data Impressions capture enough information about a robust Teacher to pass this property on to smaller Students.

%{\color{red}We posit that since adversarially trained networks are better-equipped to approximate the posterior probabilities over the adversarially perturbed data~\cite{madry-iclr-2018}, Data Impressions produced by them implicitly capture the effects of perturbations on the training distribution.} {We posit that since \emph{Data Impressions} behave as surrogates of the training data, they are also able to recover the adversarial perturbations applied to them. These \emph{DI}s can therefore help train the \emph{Student} to be robust to these perturbations in ZSKD from robust \emph{Teachers}}
We posit that since adversarially trained networks are better-equipped to approximate the posterior probabilities over the adversarially perturbed data~\cite{madry-iclr-2018}, the \textit{Data Impression} generating process is able to draw samples from the %captures the 
perturbed training distribution. In other words, the produced \textit{Data Impressions} behave as surrogates to the perturbed training data, which when used for distillation, allow the Student to also be adversarially robust.

To demonstrate this, we craft Data Impressions from adversarially-trained Teachers by exactly following the methodology described in Section~\ref{sec:proposed-approach}. Without enforcing explicit regularization or any additional penalty, we are able to produce robust Student networks under knowledge distillation in the data-free scenario.

In Table~\ref{tab:robustness-mnist}, %\ref{tab:robustness-fmnist},and~\ref{tab:robustness-cifar10}, 
we experimentally compare the performance of Student networks distilled from Data Impressions crafted from both naturally-trained and adversarially robust Teacher networks when subjected to commonly used adversarial attacks, viz., FGSM\cite{goodfellow2014explaining}, iFGSM\cite{kurakin2016adversarial}, PGD\cite{madry-iclr-2018}. The Teacher networks (as described in Section 4.1.2 for MNIST, F-MNIST, CIFAR-10) are made robust through PGD adversarial training~\cite{madry-iclr-2018}. While, it is interesting to note that the Students distilled through ZSKD from non-robust Teachers show slightly improved adversarial accuracies than the Teachers themselves, the students are not completely robust. In the case of robust Teachers however, significant robustness is passed down to the Student networks.

In subsequent sections, we present other applications to demonstrate the general applicability of Data Impressions as a surrogate to the true training data distribution when the latter is unavailable.%by virtue of their closeness to the true training data.

\begin{figure*}[htp]
\centering
 \includegraphics[width=0.95\textwidth, height=0.25\textwidth]{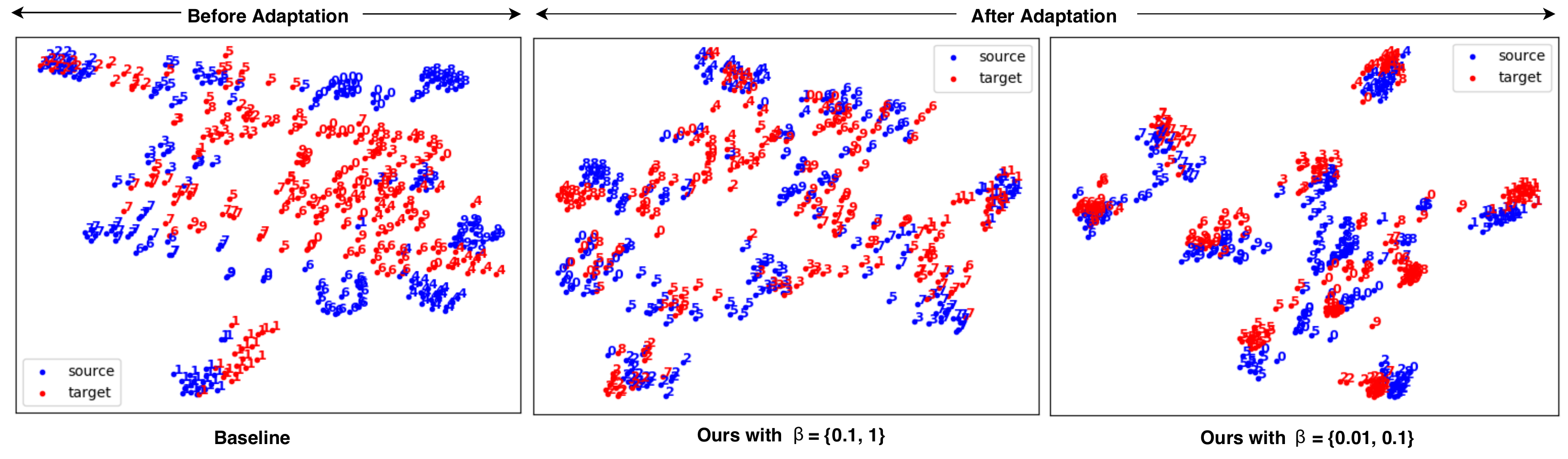}
\caption{TSNE Plots to visualize the source free domain adaptation of USPS $\rightarrow$ MNIST through our proposed approach via Data Impressions}
\label{tsne}
\end{figure*}
\vspace{-0.1in}
\subsection{Domain Adaptation}
\label{subsec:da}
%Start putting content here as another great application of DIs.
In this section, we demonstrate the applicability of Data Impressions for the task of unsupervised closed set Domain Adaptation. 

A model trained on data samples from a source distribution often does not generalize %(performs poorly) 
well when it encounters samples from a different target distribution due to domain gap or the dataset bias. In cases where the target data is unlabelled, possibility of finetuning the source model on target dataset becomes impractical. In order to reduce this domain shift, unsupervised domain adaptation techniques have gained a lot of attention recently. Based on the %relationship 
overlap between source and target label sets, there are different categories of domain adaptation: closed set, partial, open set and universal~\cite{you2019universal}. We restrict our discussion to closed set domain adaptation where the labels are shared between source and target domains.

During the deployment of source model, the source data that has been used for training may not be available due to several reasons such as data privacy, proprietary rights over the data, cost associated with sharing a large  dataset etc. (also explained in section 1). However, most of the existing works depend on the availability of both the source and target data for domain adaptation (also discussed in unsupervised domain adaptation paragraph of section~\ref{sec:related-works}). %Relatively less effort is dedicated towards unsupervised domain adaptation in the absence of the source data. Recently, Kundu~\textit{et al.}~\cite{kundu2020universal} attempted to solve this problem. However their method assumes that the source model is able to detect samples which are out of source distribution. 
We overcome this limitation by generating Data Impressions from the source model that act as a proxy to the source data. Thus, Data Impressions enable any relevant domain adaptation technique to be easily adopted for source free domain adaptation task.
%\vspace{-0.1in}
\subsubsection{Experimental Setup and Datasets}
We perform experiments on SVHN~\cite{netzer2011reading}, MNIST~\cite{lecun1998gradient} and USPS~\cite{Maji:EECS-2009-159} where we adapt:

%\paragraph{\textbf{SVHN to MNIST}} 
%{\color{blue}
\textbf{SVHN to MNIST:} In order to have a fair comparison with other works, the entire training data is used for this adaptation experiment.

%\paragraph{\textbf{MNIST to USPS}} 
%{\color{blue}
\textbf{MNIST to USPS:} We use the training protocol followed in~\cite{long2013transfer} where $2000$ and $1800$ images are sampled from MNIST and USPS respectively.

%\paragraph{\textbf{USPS to MNIST}}
%{\color{blue}
\textbf{USPS to MNIST:} We use the same training protocol as followed in the MNIST to USPS experiment.\\
All the training data are resized to $28 \times 28$ and pixel values are normalized between $0$ and $1$. We use the same LeNet architecture as described in~\cite{tzeng2017adversarial} for all the domain adaptation experiments. Note that the target data labels are not used while training. 
\subsubsection{Implementation}
We use a popular unsupervised domain adaptation technique by Tzeng \textit{et al.}~\cite{tzeng2017adversarial} as a backbone, owing to its effectiveness and simplicity. We use their implementation~\cite{adda} to get the baseline performances. Overview of our proposed method is shown in Figure~\ref{source-free-adaptation-approach}. In step $1$, the Data Impressions are generated from the pretrained source network using Algorithm~\ref{algo:zskd}. In the second step, the pretrained source model is frozen and the parameters of the target model are learned. The target model is initialized with weights of pretrained source network. The input to the source and target models are Data Impressions and unlabeled target data respectively. The outputs of the source and target models are then fed to an % Their outputs are connected to 
adversarial domain discriminator, which is trained with the objective of correctly identifying the domains of the inputs. The discriminator has two fully connected layers of $500$ neurons each with leaky ReLU as activation function and the final layer yields two outputs. %The discriminator is trained with an objective to identify the domains whereas 
The target model, however, is trained to confuse the discriminator using the adversarial loss. Finally, step $3$ performs the inference, where the trained target model is evaluated on the target data.

\begin{table}[htp]
\caption{Comparison with source dependent domain adaptation works. Notations- S:SVHN, M:MNIST, U:USPS}
\label{domain-adaptation-soa}
\centering
\begin{tabular}{|l|c|c|c|}
\hline
\pbox[c][.5cm][c]{4.0cm}{\textbf{Method} (Source $\rightarrow$ Target)}            & \textbf{S $\rightarrow$ M} & \textbf{U $\rightarrow$ M} & \textbf{M $\rightarrow$ U} \\ \hline \hline
\pbox[c][.5cm][c]{3.0cm} {Gradient Reversal\cite{ganin2016domain}}  & 0.739                                 & 0.730         & 0.771         \\ \hline
\pbox[c][.5cm][c]{3.0cm}{Domain Confusion \cite{tzeng2015simultaneous}}   & 0.681         & 0.665       & 0.791       \\ \hline
\pbox[c][.5cm][c]{2.0cm}{CoGAN \cite{liu2016coupled}}           & -                                     & 0.891       & \textbf{0.912}   \\ \hline
\pbox[c][.5cm][c]{2.0cm}{ADDA \cite{tzeng2017adversarial}}            & 0.76          & \textbf{0.901}     & 0.894       \\ \hline \hline
\pbox[c][.5cm][c]{2.0cm}{Baseline}          & 0.612          & 0.573     & 0.765       \\ \hline
%\pbox[c][.75cm][c]{2.5cm}{\textbf{Ours} (Source-free)\\($\beta=\{0.1, 1.0\}$)} & 0.690                                 & 0.7505                                & 0.888                                  \\ \hline
\pbox[c][.75cm][c]{2.5cm}{\textbf{Ours} (Source-free)\\($\beta=\{0.01, 0.1\}$)} & \textbf{0.866}                                 & 0.8915                                & 0.91 \\
\hline
\end{tabular}
\end{table}

\subsubsection{Results and Discussion}
Results are presented in Table ~\ref{domain-adaptation-soa}. The baseline performance represents directly utilizing the source model (without domain adaptation) to predict the labels for the target data. In our experiments with multiple choices for mixtures of $\beta$, we have typically observed that with lower $\beta$ values we achieve better performance. For example, with a mixture of $\beta=\{0.1, 1.0\}$, we achieve substantially better than the baseline results. However,  %We generate Data Impressions with two mixtures of $\beta$ i.e., $\{0.1, 1.0\}$ and $\{0.01, 0.1\}$. Note that the number of generated Data Impressions does not exceed that of samples in the source dataset. The performance using Data Impressions obtained via either of the two mixtures of $\beta$ results in significant improvement over the baseline. 
it can be observed from Table~\ref{domain-adaptation-soa} that $\beta$ when taken as mixture of $\{0.01, 0.1\}$ gives the best results across all the datasets. This is in line with the fact that \textit{lower $\beta$ values encourage more class specific Dirichlet softmax vectors to be sampled} (section~\ref{subsec:crafting}). 

In order to better understand, we use TSNE plots for visualization in Figure~\ref{tsne}, where USPS is adapted to MNIST.  We can observe that before adaptation, the source and target data are not aligned. After adaptation using Data Impressions, the source and target data starts getting aligned. With proper mix of $\beta$ values, the target data samples are well separated and the data clusters become more compact and tight. 

We further compare our proposed approach with other works that use source data as shown in Table~\ref{domain-adaptation-soa}. It can be easily observed that domain adaptation using Data Impressions gives competitive or better domain performance over several source dependent techniques.

\subsubsection{Comparison with Recent Source-free %Independent 
Domain Adaptation methods}
In this section, we compare our results on several datasets against some of the recent source-free domain adaptation works (refer section~\ref{sec:related-works} for the method comparison). 

\begin{table}[htp]
\caption{Comparison with SDDA. Notations- S:SVHN, M:MNIST, U:USPS}
\label{tab:sfda_comparison}
\centering
\begin{tabular}{|c|c|c|c|}
\hline
\textbf{Method} & \begin{tabular}[c]{@{}c@{}}\textbf{Source} \\ \textbf{baseline}\end{tabular} & \textbf{Adaptation} & \begin{tabular}[c]{@{}c@{}}\textbf{Improvement} \\ \textbf{over baseline}\end{tabular} \\ \hline \hline
SDDA (S $\rightarrow$ M) & 67.2  & 76.3  & 9.1            \\ \hline
\textbf{Ours} (S $\rightarrow$ M) & 61.74 & 93.55 & \textbf{31.81} \\ \hline \hline
SDDA (M $\rightarrow$ U) & 82.5  & 88.5  & 6.0            \\ \hline
\textbf{Ours} (M $\rightarrow$ U) & 83.81 & 94.56 & \textbf{10.75} \\ \hline
\end{tabular}
\end{table}
Kurmi \textit{et al.}~\cite{kurmi2021domain} proposed \textit{`Source Data free Domain Adaptation' (SDDA)} method to handle the unavailability of source data during unsupervised domain adaptation. We compare our performance on their source network architecture on two different adaptations : SVHN to MNIST and MNIST to USPS. As per their protocol, we use the entire training data of the datasets for adaptation, unlike previous experiments where only 1800 and $2000$ training images of USPS and MNIST were used. For both the adaptations, we train the network on the source data with learning rate $0.001$ and adam optimizer. The data impressions are generated with learning rate $0.001$. We use a $\beta$ mixture of $0.01$ and $0.1$ during generation of data impressions. The adaptation with generated impressions are performed with learning rate $2e^{-4}$ and adam optimizer. The results obtained are compared with SDDA as shown in Table~\ref{tab:sfda_comparison}. Our method performs significantly better and achieves a large improvement of $31.81\%$ and $10.75\%$ over baseline while performing adaptation from SVHN to MNIST and MNIST to USPS respectively.

\begin{table}[htp]
\caption{Comparison with SHOT. Notations- S:SVHN, M:MNIST, U:USPS}
\label{tab:shot_comparison}
\centering
\begin{tabular}{|c|c|c|c|}
\hline
\textbf{Method} & \begin{tabular}[c]{@{}c@{}}\textbf{Source} \\ \textbf{baseline}\end{tabular} & \textbf{Adaptation} & \begin{tabular}[c]{@{}c@{}}\textbf{Improvement} \\ \textbf{over baseline}\end{tabular} \\ \hline \hline
SHOT (S $\rightarrow$ M) & 65.72 & 89.62 & 23.9           \\ \hline
\textbf{Ours} (S $\rightarrow$ M) & 61.20 & 86.6  & \textbf{25.4}  \\ \hline\hline
SHOT (U $\rightarrow$ M) & 58.55 & 88.65 & 30.1           \\ \hline
\textbf{Ours} (U $\rightarrow$ M) & 57.30 & 89.15 & \textbf{31.85} \\ \hline \hline
SHOT (M $\rightarrow$ U) & 76.06 & 85.83 & 9.77           \\ \hline
\textbf{Ours} (M $\rightarrow$ U) & 76.50 & 91.0  & \textbf{14.5}  \\ \hline
\end{tabular}
\end{table}
Liang \textit{et al.}~\cite{liang2020we} proposed \textit{`Source HypOthesis Transfer' (SHOT)} which uses different source network architectures for adaptation for MNIST $\leftrightarrow$ USPS and SVHN $\rightarrow$ MNIST. Moreover, their proposed networks are customized with addition of batchnorm layers and weight normalization layers at the end of the feature extraction module and classifier module respectively. In order to have a fair comparison with ours, we make some modifications to the SHOT pipeline. Specifically, we replace their architectural dependent source network with our network and the "smooth loss function" used in their method is replaced with traditional cross entropy, as used for training our network. %Since we followed the protocol as mentioned in~\cite{long2013transfer}, 
Similar to ours, we use $1800$ and $2000$ images sampled from USPS and MNIST respectively while performing adaptation of the classifier from USPS to MNIST and MNIST to USPS. Also, we use the same data preprocessing as used in ours i.e. normalizing each input pixel between $0$ to $1$. %We incorporate these changes in their official code and compare the obtained results with ours in Table~\ref{tab:shot_comparison}. 
The adaptation performance achieved by SHOT on these aforementioned modifications is compared vis-a-vis ours in Table~\ref{tab:shot_comparison}. As evident, our improvement in performance over baseline is better on adaptations across different datasets.

%{\color{blue}
 In Tables~\ref{tab:sfda_comparison} and ~\ref{tab:shot_comparison}, the difference in the source baseline performances between ours and compared methods, can be attributed to the chosen hyperparameters such as initial learning rate, number of epochs, learning rate scheduler, etc. used for training the source network. The performance of SDDA mentioned in Table~\ref{tab:sfda_comparison} for different adaptations (SVHN $\rightarrow$ MNIST and MNIST $\rightarrow$ USPS) are the numbers reported from their paper. However, the weights of their pretrained source network were not available. Thus, we trained their source network architecture and performed our adaptation on it. Nevertheless, in order to have a fair comparison and to discount the performance difference in the baseline, we compare the improvement in performance over the baseline (i.e. difference between the method’s domain adaptation and its source baseline performance) between ours and SDDA methods. Similarly, in Table~\ref{tab:shot_comparison}, we reported the performance of SHOT on our architecture and then compared our performance. We used the default hyperparameters of the official github repository of SHOT while training the source network on our architecture. That resulted in better source baseline performance of SHOT (SVHN $\rightarrow$ MNIST) as compared to ours.  However, it is evident from both the Tables that we obtain more improvement in performance over the source baselines which demonstrates the efficacy of our proposed adaptation technique.%} 
\begin{figure*}[htp]
\centering
 \includegraphics[width=\textwidth, height=0.35\textwidth]{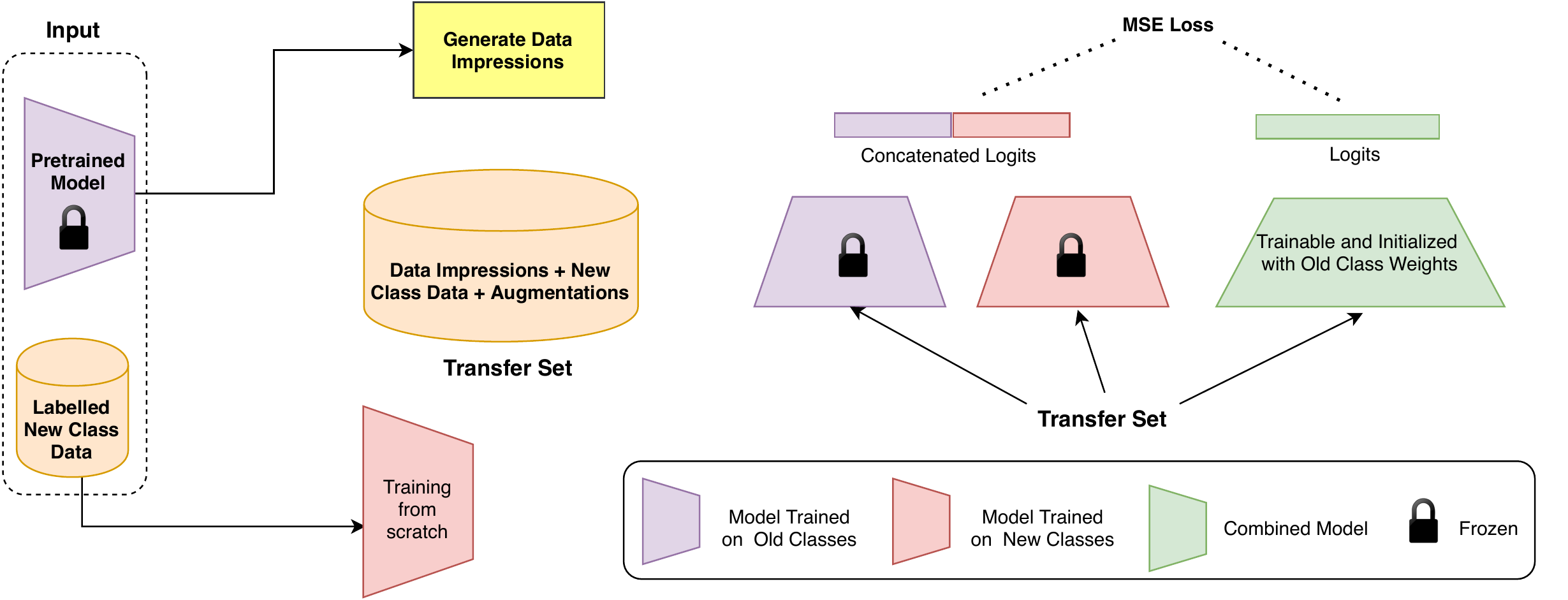}
\caption{Proposed Approach for Continual Learning using Data Impressions in the absence of old class data.}
\label{old-class-data-free-incremental-learning}
\end{figure*}
\vspace{-0.2in}
\subsection{Continual Learning}
\label{subsec:cl}
%Start putting content here as another great application of DIs.
In this section, we present the application of Data Impressions for the task of continual learning. There are several flavours of continual learning such as class incremental learning, domain incremental learning and task incremental learning \cite{van2019three}. We demonstrate the usability of Data Impressions for incrementally learning objects from different classes. In this setting, the knowlede obtained by neural network model from old classes is compromised while trying to learn from new classes. The exemplars from old classes cannot be stored due to the implicit assumption of limited memory constraints. In order to have a fair comparison, we restrict our discussion with works which do not use any exemplars from old classes. Therefore we do not consider works such as~\cite{rebuffi2017icarl, castro2018end} that store exemplars which are carefully selected to avoid catastrophic forgetting.

Since the training data that belongs to old classes is not available, some simple baselines can be adopted such as finetuning and fixed representation. In the former case, the model which is previously trained on old classes is finetuned with labelled samples of new classes while in the latter case, the model is frozen and only the last layer are trained that are connected to the new class labels. %\sout{Both of these approaches either perform well on old classes or new classes but not on both.} %Li~\textit{et al.}
LwF~\cite{li2017learning} is an important baseline that we compare against. They utilize samples from new categories for minimizing (i) the distillation loss on the old classes in order to avoid catastrophic forgetting, and (ii) cross entropy loss on the new classes. We also do comparison of our proposed method with another recent method named \textit{Deep Model Consolidation} (DMC) by Zhang~\textit{et al.}~\cite{zhang2020class}) which utilized publicly available auxiliary data for class incremental learning in the absence of exemplars. %\sout{We may not have the luxury of such unlabelled data in several image domains such as medical imaging, satellite/aerial imaging, etc.} 
Our method synthesizes data impressions using the model trained on old classes, which are then used as a substitute to the samples belonging to old categories. Hence, unlike~\cite{zhang2020class}) our proposed approach for continual learning do not require any arbitrary data.
%{\color{red} Recent work by Zhang~\textit{et al.}~\cite{zhang2020class} utilized publicly available auxiliary data for class incremental learning. However, we may not have the luxury of such unlabelled data in several image domains such as medical imaging, satellite/aerial imaging, etc. In such cases, Data Impressions can be generated using the model trained on old classes and can act as a substitute to the samples belonging to old categories.} 

\begin{figure}[htp]
\centering
 \includegraphics[width=0.49\textwidth]{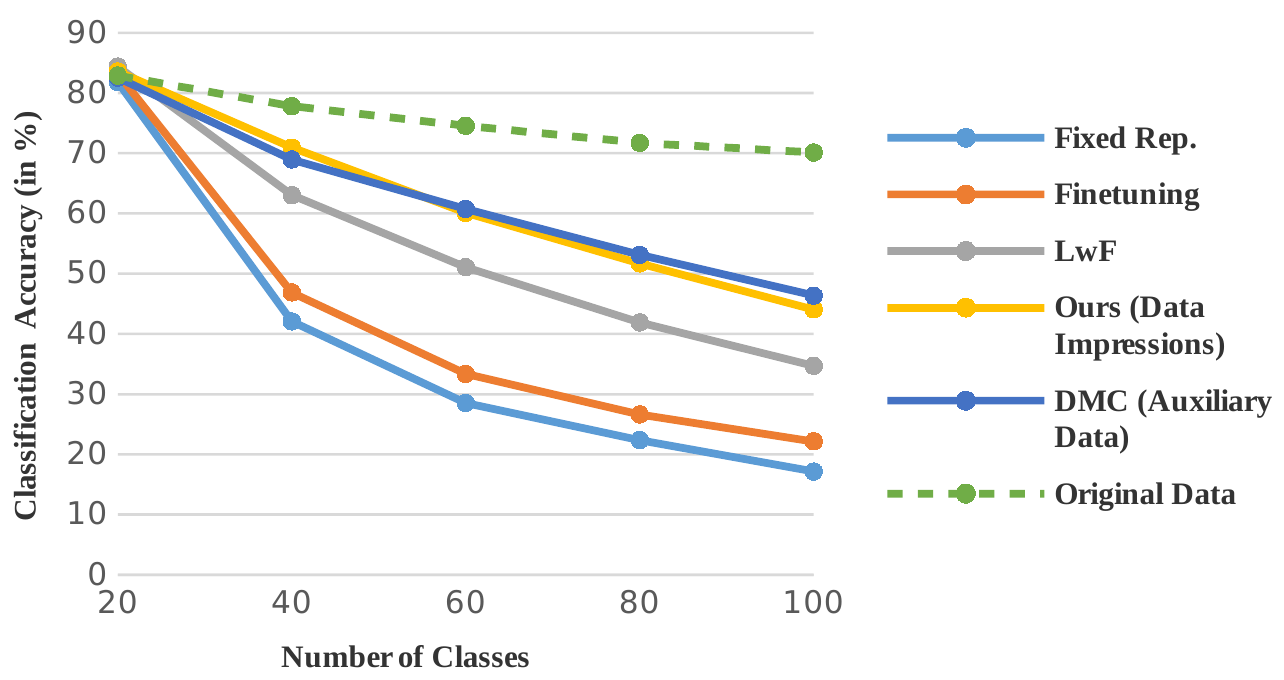}
\caption{Performance comparison of incremental learning experiments on CIFAR-100 dataset with a step size of $20$ classes.}
\label{incremental-graph}
\end{figure}

\subsubsection{Experimental Setup and Datasets}
The experiments are performed on CIFAR-100 %{\color{red}} 
dataset~\cite{krizhevsky2009learning} with an incremental step of $20$ classes. The data is normalized with channel mean and standard deviation of 0.5, then the normalized data is fed as an input to the model. In order to have a fair comparison, we use the same model architecture as in~\cite{rebuffi2017icarl, castro2018end, li2017learning, zhang2020class} i.e., the ResNet-32 \cite{he2016deep}.\\
In this exemplar-free setup, for each subsequent incremental step, apart from the samples of new classes, we only have access to the model weights trained on the old classes, but not the old class data samples themselves.
%\vspace{-0.1in}
\subsubsection{Implementation}
The proposed approach is shown in Figure~\ref{old-class-data-free-incremental-learning}. Since we consider a limited memory scenario, we generate only $2400$ Data Impressions overall. As the count of old classes increases after few incremental steps, the number of Data Impressions generated per class decreases and hence representing old classes with less generated data is challenging. Therefore, we perform simple augmentations such as flipping, rotations, scaling etc. on the generated data impressions. The dual distillation loss%as mentioned in 
~\cite{zhang2020class} is used for training the combined model. Note that unlike~\cite{zhang2020class}, we do not use any auxiliary data, instead the generated Data Impressions and labelled samples of new classes are used as a transfer set. Also, while training the combined model, we initialize with the weights of old class model as it results in better performance compared to training from scratch.

When we independently train the model on new classes data, we use an initial learning rate of $0.1$. The combined model is trained with an initial learning rate of $0.01$ for all the incremental steps except for the last incremental step where we use a learning rate of $0.001$. Across all the incremental experiments, we use SGD optimizer with momentum of $0.9$ and weight decay of $0.0005$. The learning rate is reduced by $1/5$ after every $70$ epochs and training is done for a maximum of $500$ epochs. 

\subsubsection{Results and Discussion}
The results are shown in Figure~\ref{incremental-graph} where the mean accuracy of $5$ trials is reported. We perform significantly better than {LwF}\cite{li2017learning} at every incremental step and close to {DMC}\cite {zhang2020class} (which uses additional auxiliary data). The incremental learning performance using all the samples of original data are also shown through dashed lines which serves as an upper bound.

%{\color{red} More insights here.}\\
The Fixed Representation and Finetuning baselines have severe limitations. Both of these approaches either perform well on old classes or on new classes but not on both. In the exemplar-free incremental learning setup, the challenge is to balance the performance for both the old and new classes. %, without sacrificing either. 
However, in the Fixed Representation approach, the model does not have enough capacity to learn the new classes very well and so its performance ends up being biased towards the old classes. On the other hand, in the Finetuning approach, the entire model is updated for the new classes, and so the performance is biased towards the new classes. In our approach, we generate and utilize DIs as pseudo-exemplars from the old class data and use it in conjunction with the data samples from the new classes in each incremental step. This enables achieving a nice balance in performance across both the old and new classes, as evidenced by a major improvement in performance over the aforementioned methods (see Fig.~\ref{incremental-graph}).%Our approach using DIs generates pseudo-exemplars and therefore manages to balance the performance on both the old and new classes, as we can see from the major improvement in performance reported through Figure~\ref{incremental-graph}.

Our method reports a performance very similar to DMC. However, DMC carries its own set of limitations: It utilizes auxiliary data in the absence of exemplars. Availability of such data is a strong assumption, especially considering our strictly ``data-free" experimental setting. Infact, we may not have the luxury of such unlabelled data in several specialized domains such as medical imaging, satellite/aerial imaging, etc. Furthermore, DMC's performance is dependent on how close the choice of the auxiliary data is to the original training data. Our approach overcomes these limitations
by using \textit{Data Impressions} as surrogates to the old classes which makes our method independent of any additional auxiliary data. %since \textit{Data Impressions} behave as surrogates to the old classes, %and the performance is independent of any choice of arbitrary auxiliary data.}
%and our method is also independent of arbitrary auxiliary data.}
%{\color{red} DMC: We may not have the luxury of such unlabelled data in several image domains such as medical imaging, satellite/aerial imaging, etc. Also, their perf depends on choice of arbitrary data, ours uses DIs of old classes so more reliable}
%\vspace{-0.1in}
\subsection{Universal Adversarial Perturbations}
In this section, we demonstrate the use of Data Impressions to craft Universal Adversarial Perturbations (UAPs) \cite{moosavi2017universal}. These perturbations are input-agnostic imperceptible noises that, when added to the natural data samples, can fool a target classifier into misclassifying them.

UAPs are typically powerful attacks even in the black-box setting, and it is critical to study them, especially as they have been shown to be effective in the data-free scenario. Mopuri \textit{et al.}~\cite{mopuri2018ask} realize data-free UAPs by training a generative model using Class Impressions. We craft UAPs by utilizing Data Impressions, and compare the results in Table~\ref{AAA}. 

\subsubsection{Experimental Setup and Datasets}
We use the Data Impressions  obtained from the LeNet and AlexNet classifiers described in sections \ref{subsec:mnist} (for MNIST), \ref{subsec:fmnist} (for FMNIST), and \ref{subsec:cifar} (for CIFAR-10) respectively.

We use a generator architecture modified from \cite{salimans2016improved} for a 32$\times$32 input, consisting of 4 deconvolutional layers, to generate the UAPs. The final layer is a $tanh$ nonlinearity scaled by $\epsilon$, in order to generate UAPs within the imperceptible $\epsilon$-ball. For a fair comparison, inline with the existing works, an $\epsilon$ value of $10$ is chosen for imperceptible perturbation in the $[0, 255]$ range, and is scaled accordingly with the signal range of our input.

\subsubsection{Implementation}
We train the generator that takes a mini-batch of random vectors $z$ sampled from a uniform distribution $U[-1,1]$ as input and converts them into UAPs through a series of deconvolution layers. The objective for the generator consists of a Fooling Loss and a Diversity Loss, taken from~\cite{mopuri2018ask} and used in linear combination as described therein. 

\begin{figure}[t]
\centering
\includegraphics[width=0.12\textwidth]{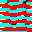}\hspace{12pt}
\includegraphics[width=0.12\textwidth]{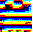}\hspace{12pt}
\includegraphics[width=0.12\textwidth]{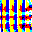}
\caption{%UAPs crafted from i) MNIST Data Impressions ii) F-MNIST Data Impressions and iii)  CIFAR-10 Data Impressions
Visualization of the UAPs crafted from CIFAR-10 Data Impressions. %that are obtained from Alexnet classifier pretrained on CIFAR-10 
}
\label{uap-di}
\end{figure}

The generator maps the latent space $Z$, consisting of 10-dimensional random vectors sampled from $U[-1,1]$ with a minibatch size of 32, to the UAPs for the target classifier. The architecture remains unchanged for all the experiments, and the generator objective is optimized using Adam. {The generator is trained for 20 epochs with a batch size of 32 for each experiment. A hyperparameter $\alpha$ is used to scale the Diversity Loss ~\cite{mopuri2018ask} before adding it to the Fooling Loss. For CIFAR-10, an $\alpha$ of 3e-04, and a learning rate of 1e-05 is used. For both FMNIST and MNIST however, an $\alpha$ of 1e-04 and a learning rate of 1e-05 is used}. Figure~\ref{uap-di} shows sample UAPs learned by using Data Impressions extracted from target classifier (Alexnet) pretrained on CIFAR-10.

\subsubsection{Results and Discussion}
Table \ref{AAA} compares the fooling rates of the UAPs crafted from Data Impressions and Class Impressions crafted from the same classifiers. It can be observed that the UAPs from Data Impressions achieve better fooling rates and outperform those of Class Impressions by a minimum of $4.05\%$. Owing to the better representation of the underlying training data by our Data Impressions compared to the Class Impressions~\cite{aaa-eccv-2018}, the same generative model can now craft better perturbations which is manifested in the form of better fooling rates. 

The class impressions are estimated as inputs that maximize the softmax outputs/logit, corresponding to the specific class. Therefore, it is obvious that the CIs are class-specific and the samples generated for each class exhibit very little diversity. On the contrary, the DIs we estimate are not tied to any specific class and are generated for softmax vectors sampled from a Dirichlet distribution with diverse values of the entropy of the target softmax. This leads to the possibility of creating a training set for UAP generation, composed of statistically uncorrelated as well as visually diverse image samples. In fact, the CIs can be shown to be DIs generated for one-hot encoded target softmax vectors, thereby making them just a special case and a small subset of the corresponding set of data impressions. Due to this improvement in quality of the image set, we are able to craft diverse set of strong UAPs leading to better fooling rates.%The class impressions tries to maximize the outputs/logits which is intuitively and theoretically similar to generating samples to match a one-hot encoded target output. On the other hand, we not only generate samples for targets which are similar to that in terms of entropy but we also generate samples for higher entropy target distribution as sampled from the dirichlet. This not only increases the size of uncorrelated samples in the training set for generating UAPs but also increases the visual diversity as well. That's why we achieve better fooling rates while using data impressions for crafting UAPs.

%{\color{red}More insights?}
\begin{table}[htp]
\caption{Comparison of Fooling Rates (in \%) of UAPs crafted from Class Impressions and Data Impressions}
\label{AAA}
\centering
\begin{tabular}{|c|c|c|c|l}
\cline{1-4}
Method & \begin{tabular}[c]{@{}c@{}}AlexNet \\ (CIFAR-10)\end{tabular} & \begin{tabular}[c]{@{}c@{}}LeNet\\ (Fashion-MNIST)\end{tabular} & \begin{tabular}[c]{@{}c@{}}LeNet\\ (MNIST)\end{tabular} &  \\ \cline{1-4}
CI: AAA\cite{mopuri2018ask} & 90.18 & 91.29 & 91.10 &  \\ \cline{1-4}
DI: Ours & \textbf{94.23} & \textbf{96.37} & \textbf{96.45} &  \\ \cline{1-4}
\end{tabular}
\end{table}
\vspace{-0.1in}
\section{Key Observations}
\label{sec:overall_discussion}
%{\color{red}Summary of all the results, discussions and insights from the various applications}

%{\color{red} Extracting surrogates for the training data from the parameters of a trained neural network is a nontrivial task. }
In this section, we summarize our major findings based on the extensive experiments performed with data impressions across multiple different applications.

In ZSKD, $\beta$ is an important scaling parameter which controls the spread of the Dirichlet distribution. Empirically, we observed better performance when $\beta$ is a mixture of {$1.0$ and $0.1$}. This encourages higher diversity (variance) and at the same time does not result in highly sparse vectors in comparison to the smaller $\beta$ mixture of {$0.1$ and $0.01$}. Robustness of the Teacher implicitly gets transferred to the Student models through the data impressions, without explicitly adding any extra objective during its generation. Thus, our proposed method for extracting impressions by design itself, closely approximates the data distribution on which the Teacher network is trained. 

Another interesting observation is that the student models distilled using data impressions from a non-robust teacher network, obtains slightly higher adversarial accuracy across several adversarial attacks over different datasets in comparison to the performance of corresponding Teacher. This robustness in Student networks can be explained with the fact that the data impressions do not visually `look' exactly like the training images themselves, but actually only capture the `essence' of the training data. Thus generated synthetic data impressions are analogous to that of `adversarial' samples with no bound on perturbation (no $\epsilon$ constraint) as the Teacher network classifies them similar to original training samples. 

Class impressions~\cite{aaa-eccv-2018} can be considered as a special case of data impressions. Small values of $\beta's$ are chosen to enforce the softmax vectors to be sampled from the corners of the simplex in the Dirichlet distribution, making them highly class specific (Proof provided in the supplementary). Based on the experiments performed across multiple datasets, it is evident that the data impressions have clearly outperformed class impressions in both distillation and also in crafting the UAPs. Hence, modelling the softmax space via Dirichlet distribution for extracting surrogate samples is better in comparison to the one-hot vector modelling.

For domain adaptation, the data impressions generated with smaller $\beta's$ (e.g. mixture of $0.1$ and $0.01$) works better. This shows that the diversity induced in the impressions through high $\beta$ is not as important as retaining more class information with lower $\beta's$ for this application. In the case of incremental learning, we performed distillation from two separate models trained with old and new classes data respectively into a combined model. We used the transfer set consisting of data impressions and new class data and observed performance as good as the DMC method~\cite{zhang2020class} which assumes access to the auxiliary data. We also observed that the initialization of the combined model with old class weights is better than training the combined model from scratch since the optimization gets easier and leads to better performance.

We choose to show the efficacy of data impressions on some of the most popular applications. We followed the benchmark problem setup and datasets to evaluate the performance of the generated data impressions. Note that these impressions are not generated specifically targeting any particular application, which makes them independent of the target application and hence they can be used in other applications beyond the ones we have demonstrated.

\vspace{-0.1in}
\section{Conclusion}
\label{sec:concluion}
In this paper we introduced a novel and interesting problem of restoring training data from a trained deep neural network. Utilizing only the parameters of the trained model but no additional prior to achieve this makes it a challenging task. Hence, we rather focused on a simplified problem. We aimed to restore the training data in a \textit{learning} sense. In other words, our objective is to restore data that can train models on related tasks and generalize well onto the natural data. Apart from the natural academic interest, the presented task has wide practical applicability. Especially it has great value in adapting the laboratory trained deep models into complex data-free scenarios as detailed in section~\ref{sec:introduction}. In that regard, we have demonstrated the fidelity of the extracted samples, known as Data Impressions, via realizing excellent generalization for multiple tasks such as Knowledge distillation, crafting Adversarial Perturbations, Incremental Learning, and Domain Adaption. However, one can notice that, although Data Impressions capture some of the striking visual patterns from the actual training data samples, they are visually far away from the training data. Strong priors about the natural training distribution might be needed to improve the visual similarity, an aspect we leave for future investigation.

\appendices

%\section{}

\ifCLASSOPTIONcompsoc
  \section*{Acknowledgments}
  This work is partially supported by 1. Start-up Research Grant (SRG) from SERB, DST, India (Project file number: SRG/2019/001938) and 2. Young Scientist Research Award (Sanction no. 59/20/11/2020-BRNS) from DAE-BRNS, India. We would like to extend our gratitude to all the reviewers for their valuable suggestions.
\else
  \section*{Acknowledgment}

\fi


\ifCLASSOPTIONcaptionsoff
  \newpage
\fi



\bibliographystyle{IEEEtran}
\normalem
\bibliography{references.bib}
\ULforem
\newpage
\onecolumn
\begin{center}
    \Large{\textbf{\textit{Supplementary for\\} ``Mining \textit{Data Impressions} from Deep Models as Substitute for the Unavailable Training Data"} }
\end{center}

\setcounter{section}{0}
\setcounter{table}{0}
\setcounter{figure}{0}
\setcounter{equation}{0}
\vspace{8pt}
\hrule
\vspace{18pt}
\section{Class Impressions - A Special Case of Data Impressions}
\label{ci_special_case_proof}
\vspace{8pt}
Class impressions by Mopuri \textit{et al.}~\cite{mopuri2018ask} are generated via logit maximization for the target class. On other hand, our proposed method synthesizes data impressions for a target category by matching the temperature raised softmax predictions of the \Te{} network with the softmax sampled from Dirichlet distribution. Hence, our impressions are optimized using cross entropy between these softmax vectors. A special case in this context is when the target softmax output is a one hot vector. We aim to show that the synthesis of class impressions through logit maximization is same as minimizing the cross entropy loss with target softmax output as one hot vector. Therefore this shows that the data impressions are generic where the target vectors can have high or low entropy by suitably adjusting the $\beta$ values whereas class impressions are a special case with target vectors of low entropy. \bigskip
%\vspace{8pt}
%\newline
%\newline
\\
\textbf{Notation:} \\
$T$ : \Te{} network\\
$x$ : Random Input\\
$T(x)$ : pre-softmax values of the trained \Te{} network for input $x$\\
$c$ : Target category \\
$T(x)^{c}$ : Logit value for a target category `c' \\
$K$ : Dimension of the output probability vector \\
$CE$ : cross entropy \\
$\beta$ : scaling factor \\
$CS^{c}$ : $c^{th}$ row of class similarity matrix $CS$ \\
$\bm{\alpha}^{c}$ : Concentration parameter for category `c' \\
$I(c) = (0, \ldots, 0, 1, 0, \ldots, 0)$ is the
one-hot vector with $1$ for class `c' and $0$ elsewhere \\
$CI$ : class impression\\
$DI$ : data impression\\
$Dir$ : dirichlet function
\bigskip
\\
\textbf{Proof:}\\
Synthesis of CI for a category $c \in {1 \ldots K}$\\ \newline
$\implies$ $\max_{x} T(x)^{c}$ \\ \newline
$\implies$ $\max_{x} softmax(T(x))^{c}$  \newline [$\because$ softmax is monotonic function]\\ \newline
$\implies$ $\max_{x} 1 \cdot \log(softmax(T(x))^{c})$ \\ \newline
$\implies$ $\min_{x} CE(softmax(T(x), I(c))$ \\ \newline
$\approx$ $\min_{x} CE(softmax(T(x), Dir(K, \beta \cdot \bm{\alpha}^{c})$ such that $\beta \ll 1$\\ \newline
$\implies$ $\min_{x} CE(softmax(T(x), Dir(K, \beta \cdot CS^{c})$ \newline such that $\beta \ll 1$\\ \newline
$\implies$ Synthesis of DI for the category $c$ with $\beta \ll 1$
\newline
\newline
Hence, CI is a special case of DI.

\newpage
\section{Comparison: Our Generated Class Similarity matrix v/s Similarity matrix computed using Real Unseen Dataset}
\vspace{8pt}
We compute the class similarity matrix using two real unseen dataset. More specifically, we perform experiments with test data of cifar-$10$~\cite{krizhevsky2009learning} and arbitrary data sharing the same category i.e. SVHN~\cite{netzer2011reading}. The data is first passed to the teacher model and the features are obtained from the pre-softmax layer. Then, we perform L2 normalization on the features. We use the labels from the teacher’s prediction. The features from a particular class are grouped together. Then we take the mean of the features that belong to a particular class and thus, we get the mean representative normalized feature for each class. 
 \\
 
We obtain a class similarity matrix $C$ where the entry in the $i^{th}$ row and $j^{th}$ column denoted by $C_{ij}$ is the similarity score computed as:
\begin{equation}
    C_{ij} = \bm{mf}_i^T\bm{mf}_j
\end{equation}
where $\bm{mf}_i$ and $\bm{mf}_j$ are the mean of normalized features for class $i$ and class $j$ respectively.
\\

Finally, class similarity matrix $C$ is normalized through min-max normalization. This class similarity matrix is compared with our generated class similarity matrix obtained using the last layer weights of the teacher network in absence of training data as mentioned in %section~\ref{subsec:dirichlet-modelling} (equation~\ref{eqn:class-sim-mat}) 
section $3.1$ (equation $1$) in the main draft. The comparison is done via calculating the Pearson and Spearman correlations between them. We perform the class similarity experiments on Alexnet teacher trained on cifar-$10$ and the results are presented below:
\begin{table}[htp]
\centering
\caption{Correlation analysis on the class similarity matrix computed by unseen test samples of cifar-10 with respect to our generated similarity matrix. 
}
\label{tab:cifar_class_similarity_correlation}
\begin{tabular}{|c|c|c|}
\hline
\textbf{Correlation Name} & \textbf{Correlation Value} & \textbf{p-value} \\ \hline\hline
Pearson                   & 0.9490                     & 5.8798  e-51     \\ \hline
Spearman                  & 0.6211                     & 5.4070 e-12      \\ \hline
\end{tabular}

\end{table}

\begin{table}[htp]
\centering
\caption{Correlation analysis on the class similarity matrix computed by unseen SVHN samples with respect to our generated similarity matrix. The teacher network’s predictions are used as labels for the test samples. }
\label{tab:svhn_class_similarity_correlation}
\begin{tabular}{|c|c|c|}
\hline
\textbf{Correlation Name} & \textbf{Correlation Value} & \textbf{p-value} \\ \hline \hline
Pearson                   & 0.7724                     & 4.8992 e-21      \\ \hline
Spearman                  & 0.5577                     & 1.6640 e-09      \\ \hline
\end{tabular}

\end{table}
As the pvalue is less than $0.05$, thus the class similarity computed using unseen data (cifar-$10$/svhn) and our proposed approach is strong positively correlated on Pearson correlation and moderate positively correlated on Spearman correlation.

%\bibliographystyle{IEEEtran}
\normalem
%\bibliography{references.bib}
\ULforem

\end{document}